\documentclass[journal]{IEEEtran}
\usepackage[numbers,sort&compress]{natbib}

\ifCLASSINFOpdf
\else
\fi
\usepackage{graphicx}
\usepackage{makecell}
\usepackage{booktabs}
\usepackage[table]{xcolor}
\graphicspath{{Figure/}}
\begin{document}
%
\title{Deep Learning for LiDAR Point Clouds \\ in Autonomous Driving: A Review}
%
%
%

\author{Ying~Li,
        Lingfei~Ma,~\IEEEmembership{Student Member,~IEEE,}
        Zilong~Zhong,~\IEEEmembership{Student Member,~IEEE,}
        Fei~Liu,
        Dongpu~Cao,~\IEEEmembership{Senior Member,~IEEE,}
        Jonathan~Li,~\IEEEmembership{Senior Member,~IEEE,}
        and~Michael~A.~Chapman~\IEEEmembership{Senior Member,~IEEE,}
\thanks{Y.Li, L.Ma and J.Li are with the Department of Geography and Environmental Management, University of Waterloo, 200
University Avenue West, Waterloo, N2L 3G1, Canada (e-mail: y2424li@uwaterloo.ca, l53ma@uwaterloo.ca).}
\thanks{Z.Zhong is with School of Data and Computer Science， Sun Yat-Sen University, Guangzhou, China, 510006 (email: zlzhong@ieee.org).}
\thanks{F.Liu is with Xilinx Technology Beijing Limited, Beijing, China, 100083 (email: fledaliu@xilinx.com).}
\thanks{D.Cao is with Waterloo Cognitive Autonomous Driving Lab, University of Waterloo,  N2L 3G1, Canada (e-mail: dongpu.cao@uwaterloo.ca).}
\thanks{J. Li is with the Departments of System Design Engineering, University of Waterloo, Waterloo, ON N2L 3G1, Canada (e-mail: junli@uwaterloo.ca).}
\thanks{M. A. Chapman is with the Department of Civil Engineering, Ryerson University, Toronto, ON M5B 2K3, Canada (e-mail:,mchapman@ryerson.ca).}}

%
%

\markboth{Journal of \LaTeX\ Class Files}%
{Li \MakeLowercase{\textit{et al.}}: Deep Learning for LiDAR Point Clouds in Autonomous Driving: A Review}
%



\maketitle

\begin{abstract}
Recently, the advancement of deep learning in discriminative feature learning from 3D LiDAR data has led to rapid development in the field of autonomous driving. However, automated processing uneven, unstructured, noisy, and massive 3D point clouds is a challenging and tedious task. In this paper, we provide a systematic review of existing compelling deep learning architectures applied in LiDAR point clouds, detailing for specific tasks in autonomous driving such as segmentation, detection, and classification. Although several published research papers focus on specific topics in computer vision for autonomous vehicles, to date, no general survey on deep learning applied in LiDAR point clouds for autonomous vehicles exists. Thus, the goal of this paper is to narrow the gap in this topic. More than $140$ key contributions in the recent five years are summarized in this survey, including the milestone 3D deep architectures, the remarkable deep learning applications in 3D semantic segmentation, object detection, and classification; specific datasets, evaluation metrics, and the state of the art performance. Finally, we conclude the remaining challenges and future researches.
\end{abstract}


%
\IEEEpeerreviewmaketitle

\section{Introduction}
%
%
%
%

\IEEEPARstart{A}{ccurate} environment perception and precise localization are crucial requirements for reliable navigation, information decision and safely driving of autonomous vehicles (AVs) in complex dynamic environments\cite{janai2017computer,levinson2011towards}. These two tasks need to acquire and process highly-accurate and information-rich data of real-world environments \cite{van2018autonomous}. To obtain such data, multiple sensors such as LiDAR and digital cameras \cite{huang2018apolloscape} are equipped on AVs or mapping vehicles to collect and extract target context. Traditionally, image data captured by the digital camera, featured with 2D appearance-based representation, low cost, and high efficiency, is the most commonly used data in perception tasks \cite{vivacqua2018self}. However, image data lack of 3D geo-referenced information \cite{remondino2011heritage}. Thus, the dense, geo-referenced, and accurate 3D point cloud data collected by LiDAR are exploited. Besides, LiDAR is not sensitive to the variations of lighting conditions and can work under day and night, even with glare and shadows \cite{wu2018squeezeseg}.

\begin{figure}[!t]
\centering
\includegraphics[width=3.5in]{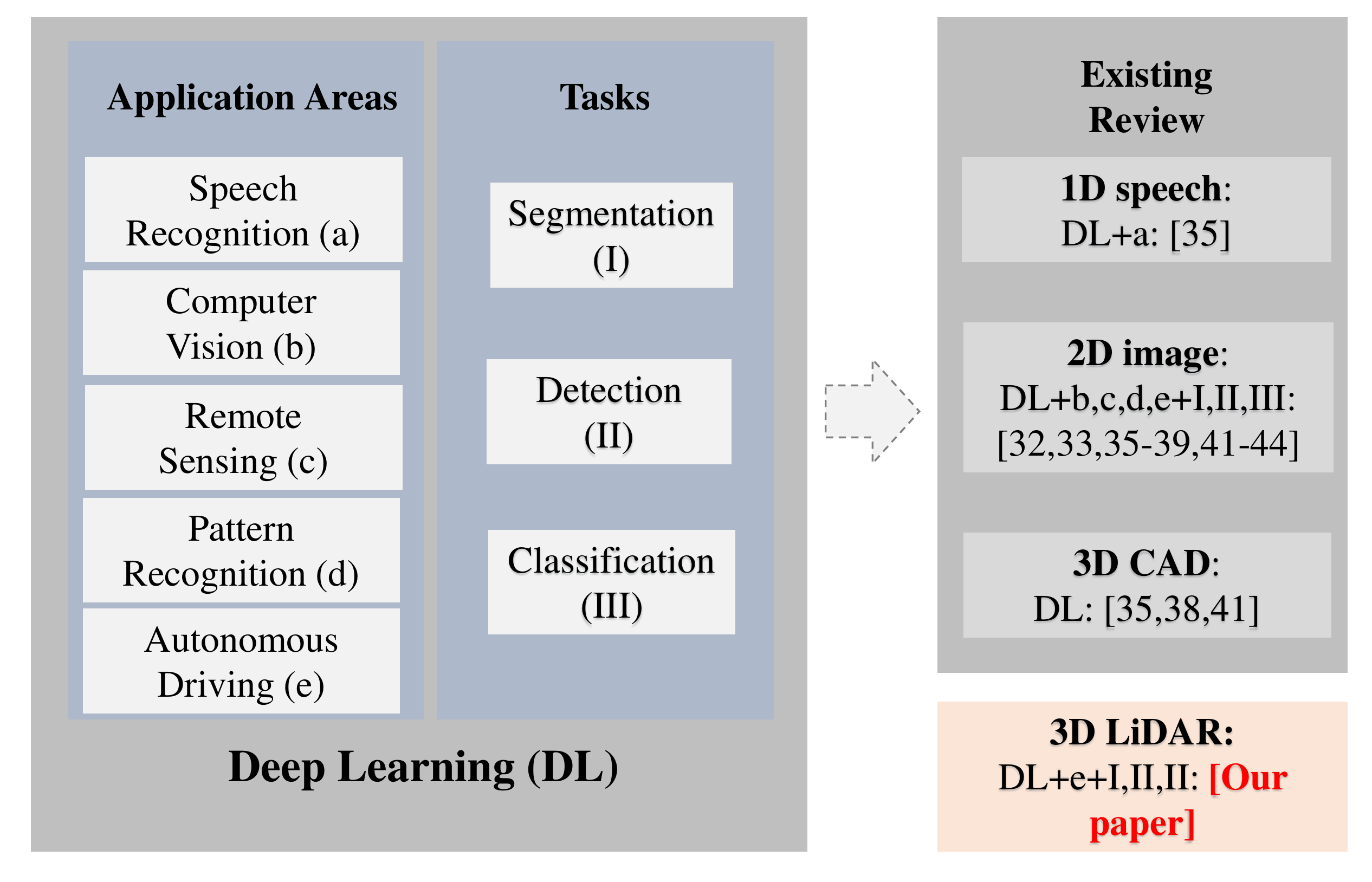}
\caption{Existing review paper related to DL and their application with different tasks. We summarize that our paper is the first one to survey the application of LiDAR point clouds in segmentation, detection and classification tasks for autonomous driving using DL techniques}\label{top}
\end{figure}

The application of LiDAR point clouds for AVs can be described in two aspects: (1) real-time environment perception and processing for scene understanding and object detection \cite{yang2018pixor}; (2) high-definition (HD) maps and urban models generation and construction for reliable localization and referencing \cite{levinson2011towards}. These applications have some similar tasks, which can be roughly divided into three types: 3D point cloud segmentation, 3D object detection and localization, and 3D object classification and recognition. Such a technique has led to an increasing and urgent requirement for automatic analysis of 3D point clouds \cite{li2016vehicle} for AVs.

Driven by the breakthroughs brought by deep learning (DL) techniques and the accessibility of 3D point cloud, the 3D DL frameworks have been investigated based on the extension of 2D DL architectures to 3D data with a notable string of empirical successes. These frameworks can be applied to several tasks specifically for AVs such as: segmentation and scene understanding \cite{qi2017pointnet,boulch2017unstructured,qi2017pointnet++}, object detection \cite{zhou2018voxelnet,li20173d}, and classification \cite{qi2017pointnet,qi2016volumetric,dewan2017deep}. Thus, we provide a systematic survey in this paper, which focuses explicitly on framing the LiDAR point clouds in segmentation, detection, and classification tasks for autonomous driving using DL techniques.

Several related surveys based on DL have been published in recent years. The basic and comprehensive knowledge of DL is described in detail in \cite{lecun2015deep,sze2017efficient}. These surveys normally focused on reviewing DL applications in visual data \cite{guo2016deep,Voulodimos2018Deep} and remote sensing imagery \cite{zhang2016deep,zhu2017deep}. Some are targeted at more specific tasks such as object detection \cite{liu2018deep,zhao2019object}, semantic segmentation \cite{garcia2017review}, recognition \cite{liu2017survey}. Although DL in 3D data has been surveyed in \cite{bronstein2017geometric,ioannidou2017deep,ahmed2018deep}, these 3D data are mainly 3D CAD models \cite{wu20153d}. In \cite{janai2017computer}, challenges, datasets, and  methods in computer vision for AVs are reviewed. However, DL applications in LiDAR point cloud data have not been comprehensively reviewed and analyzed. We summarize these surveys related to DL in Fig.\ref{top}.

There also have several surveys published for LiDAR point clouds. In \cite{ma2018mobile,guan2016use,che2019object, wang2018lidar}, 3D road object segmentation, detection, and classification from mobile LiDAR point clouds are introduced, but they are focusing on general methods not specific for DL models. In \cite{hana2018comprehensive}, comprehensive 3D descriptors are analyzed. In \cite{arnold2019survey,liu2019deep}, approaches of 3D object detection applied for autonomous driving are concluded. However, DL models applied in these tasks have not been comprehensively analyzed. Thus, the goal of this paper is to provide a systematic review of DL using LiDAR point clouds in the field of autonomous driving for specific tasks such as segmentation, detection/localization, and classification.

The main contributions of our work can be summarized as:
\begin{itemize}
  \item An in-depth and organized survey of the milestone 3D deep models and a comprehensive survey of DL methods aimed at tasks such as segmentation, object detection/localization, and classification/recognition in AVs, their origins, and their contributions.
  \item A comprehensive survey of existing LiDAR datasets that can be exploited in training DL models for AVs.
  \item A detailed introduction for quantitative evaluation metrics and performance comparison for segmentation, detection, and classification. 
  \item A list of the remaining challenges and future researches that help to advance the development of DL in the field of autonomous driving.
\end{itemize}

The remainder of this paper is organized as follows: Tasks in autonomous driving and the challenges of DL using LiDAR point cloud data are introduced in Section II. A summary of existing LiDAR point clouds datasets and evaluation metrics are described in Section III. Then the milestone 3D deep models with four data representations of LiDAR point clouds are described in Section IV. The DL applications in segmentation, object detection/localization, and classification/recognition for AVs based on LiDAR point clouds are reviewed and discussed in Section V. Section VI proposes a list of the remaining challenges for future researches. We finally conclude the paper in Section VII.

\section{Tasks and Challenges}
\subsection{Tasks}
In the perception module of autonomous vehicles, semantic segmentation, object detection, object localization, and classification/recognition constitute the foundation for reliable navigation and accurate decision \cite{treml2016speeding}. These tasks are described as follows respectively: 

\begin{itemize}
\item \textbf{3D point cloud semantic segmentation}: Point cloud segmentation is the process to cluster the input data into several homogeneous regions, where points in the same region have the identical attributes \cite{nguyen20133d}. Each input point is predicted with a semantic label, such as ground, tree, building. The task can be concluded as: given a set of ordered 3D points $X= \left\{ x_1,x_2,x_i,\cdots,x_n \right\}$ with ${x_i}\in{R^3}$ and a candidate label set $Y= \left\{ y_1,y_2,\cdots,{y_k} \right\}$, assign each input point $x_i$ with one of the k semantic labels \cite{huang2018recurrent}. Segmentation results can further support object detection and classification, as shown in Fig.\ref{f_challenges}(a).
\item \textbf{3D object detection/localization}: Given an arbitrary point cloud data, the goal of 3D object detection is to detect and locate the instances of predefined categories (e.g., cars, pedestrians, and cyclists, as shown in Fig.\ref{f_challenges}(b)), and return their geometric 3D location, orientation and semantic instance label \cite{qi2018frustum}. Such information can be represented coarsely using a 3D bounding box which is tightly bounding the detected object \cite{qi2019votenet,qi2019votenet,zhou2018voxelnet}. This box is commonly represented as $(x,y,z,h,w,l,\theta, c)$, where $(x,y,z)$ denotes the object (bounding box) center position, $(h,w,l)$ represents the bounding box size with width, length and height, and $\theta$ is the object orientation. The orientation refers to the rigid transformation that aligns the detected object to its instance in the scene, which are the translations in each of the of x, y, and z directions as well as a rotation about each of these three axes \cite{beltran2018birdnet,kundu20183d}. $c$ represents the semantic label of this bounding box (object). 
\item \textbf{3D object classification/recognition}: Given several groups of point clouds, the objectiveness of classification /recognition is to determine the category (e.g., mug, table, or car, as shown in Fig.\ref{f_challenges}(c)) the group points belong to. The problem of 3D object classification can be defined as: given a set of 3D ordered points $ X= \left\{ x_1,x_2,x_i,\cdots,x_n \right\}$ with $x_i\in{R^3}$ and a candidate label set $ Y= \left\{ y_1,y_2,\cdots,{y_k} \right\} $, assign the whole point set $X$ with one of the $k$ labels \cite{luo2019learning}.
\end{itemize}

\subsection{Challenges and Problems}
\begin{figure}[!t]
\centering
\includegraphics[scale=0.45]{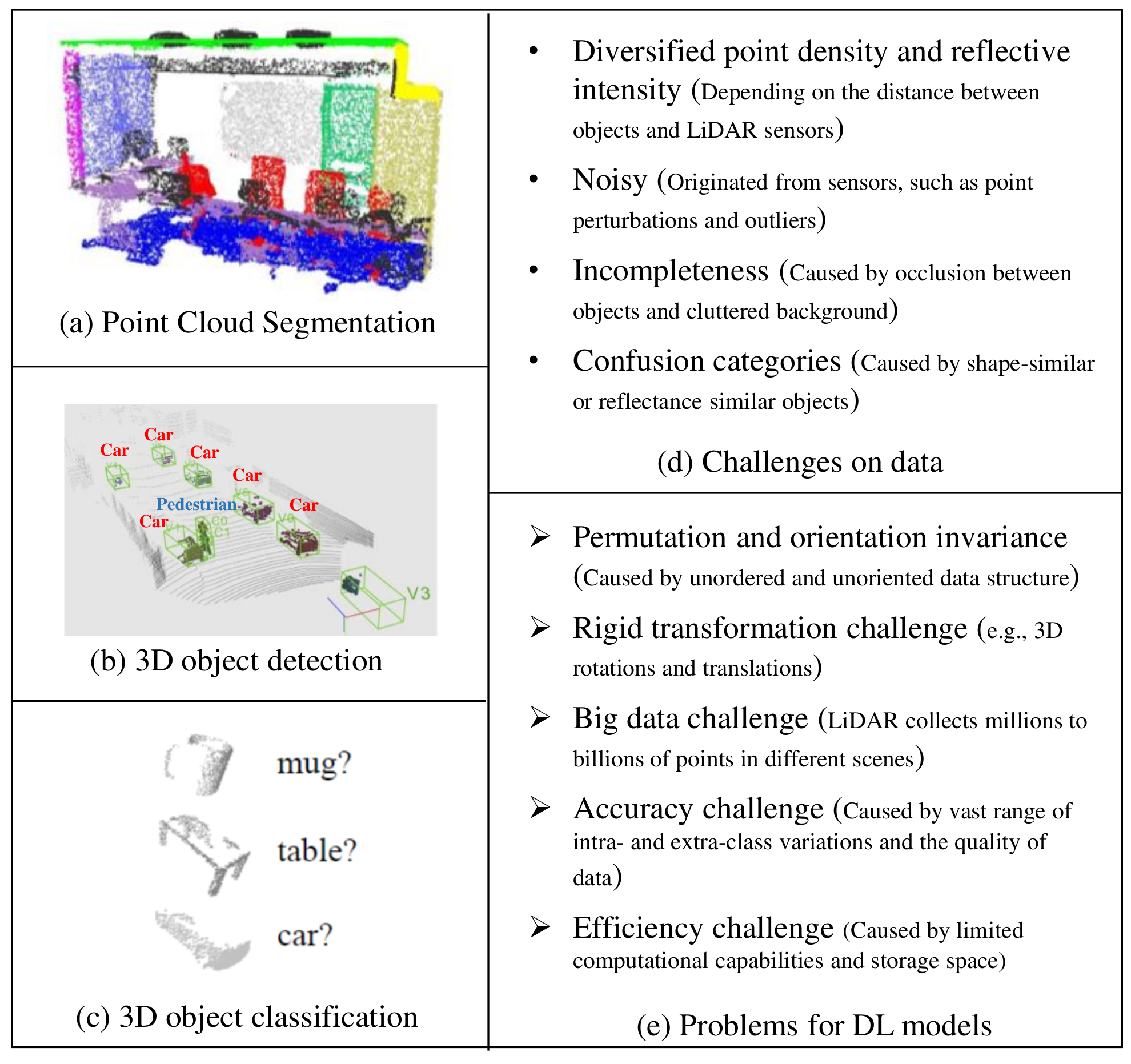}
\caption{Tasks and challenges related to DL-based applications on 3D point clouds: (a) Point cloud segmentation \cite{qi2017pointnet}, (b) 3D object detection \cite{qi2018frustum}, (c) 3D object classification \cite{qi2017pointnet}, (d) challenges on LiDAR point clouds, (e) Problems for DL models}
\label{f_challenges}
\end{figure}

In order to segment, detect, and classify the general objects using DL for AVs with robust and discriminative performance, several challenges and problems that must be addressed, as shown in Fig.\ref{f_challenges}. The variation of sensing conditions and unconstrained environments results in the challenges on data. The irregular data format and requirements for both accuracy and efficiency pose the problems that DL models need to solve.

\subsubsection{Challenges on LiDAR point clouds}
Changes in sensing conditions and unconstrained environments have dramatic impacts on object appearance. In particular, the objects captured at different scenes or instances exist a set of variations. Even for the same scene, the scanning times, locations, weather conditions, sensor types, sensing distances and backgrounds are all brought about intra-class differences. All these conditions produce significant variations for both intra- and extra-class objects in LiDAR point cloud data:
\begin{itemize}
\item \textbf{Diversified point density and reflective intensity}. Due to the scanning mode of LiDAR, the density and intensity for objects vary a lot. The distribution of these two characteristics highly depends on the distance between objects and LiDAR sensors \cite{wang2015multiscale,wen2019deep,hackel2017joint}. Besides, the ability of the LiDAR sensors, the time constraints of scanning and needed resolution also affect their distribution and intensity. 
\item \textbf{Noisy}. All sensors are noisy. There are a few types of noise that include point perturbations and outliers \cite{kumar2019multi}. It means that a point has some probability to be within a sphere of a certain radius around the place it was sampled (perturbations), or it may appear in a random position in space \cite{pang20163d}.
\item \textbf{Incompleteness}. Point cloud data obtained by LiDAR are commonly incomplete \cite{tagliasacchi2009curve}. This mainly results from the occlusion between objects \cite{pang20163d}, cluttered background in urban scenes \cite{kumar2019multi,wang2015multiscale}, and unsatisfactory material surface reflectivity. Such problems are severe in real-time capturing of moving objects, which exist large gaping holes and severe under-sampling. 
\item \textbf{Confusion categories}. In a natural environment, shape-similar or reflectance similar objects have interference in object detection and classification. For example, some manmade objects such as commercial billboards have similar shapes and reflectance with traffic signs.
\end{itemize}

\subsubsection{Problems for 3D DL models}
The irregular data format and the requirements for accuracy and efficiency from tasks bring some new challenges for DL models. A discriminate and general-purpose 3D DL model should solve the following problems when designing and constructing its framework:
\begin{itemize}
\item \textbf{Permutation and orientation invariance}. Compared with 2D grid pixels, the LiDAR point clouds are a set of points with irregular order and no specific orientation \cite{liu2019relation}. Within the same group of $N$ points, the network should feed N! permutations in an order to be invariant. Besides, the orientation of point sets is missing, which poses a great challenge for object pattern recognition \cite{huang2009consolidation}.   
\item \textbf{Rigid transformation challenge}. There exist various rigid transformations among point sets, such as 3D rotations and 3D translations. These transformations should not affect the performance of networks \cite{qi2017pointnet++,liu2019relation}.  
\item \textbf{Big data challenge}. LiDAR collects millions to billions of points in different urban or rural environments with nature scenes \cite{kumar2019multi}. For example, in Kitti dataset \cite{geiger2013vision}, each frame captured by 3D Velodyne laser scanners contains 100k points. The smallest collected scene has 114 frames, which has more than 10 million points. Such amounts of data bring difficulties in data storage.
\item \textbf{Accuracy challenge}. Accurate perception of road objects is crucial for AVs. However, the variation for both intra-class and extra-class objects and the quality of data pose challenges for accuracy. For example, objects in the same category have a set of different instances, in terms of various material, shape, and size. Besides, the model should be robust to the unevenly distributed, sparse, and missing data. 
\item \textbf{Efficiency challenge}. Compared with 2D images, processing a large quantity number of point clouds produces high computation complexity and time costs. Besides, the computation devices on AVs have limited computational capabilities and storage space \cite{jo2015development}. Thus, an efficient and scalable deep network model is critical.
\end{itemize}

\section{Datasets and Evaluation Metrics}

\subsection{Datasets}
Datasets pave the way towards the rapid development of 3D data application and exploitation using DL networks. There are two roles of reliable datasets: one for providing a comparison for competing algorithms, another for pushing the fields towards more complex and challenging tasks \cite{liu2018deep}. With the increasing application of LiDAR in multiple fields, such as autonomous driving, remote sensing, photogrammetry, there is a rise of large scale datasets with more than millions of points. These datasets accelerate the crucial breakthroughs and unpredicted performance in point cloud segmentation, 3D object detection, and classification. Apart from the mobile LiDAR data, some discriminative datasets \cite{hackel2017semantic3d} acquired by terrestrial laser scanning (TLS) by static LiDAR are also employed due to they provide high-quality point cloud data.

\begin{table*}[!ht]
\centering
\caption{Survey of existing LiDAR dataset}
\begin{tabular}{|c|c|c|c|c|c|c|}
\hline
\textbf{Dataset} & \textbf{Format} & \textbf{\begin{tabular}[c]{@{}c@{}}Primary \\ Fields\end{tabular}} & \textbf{\begin{tabular}[c]{@{}c@{}}Points /\\ Objects\end{tabular}} & \textbf{\# Classes} & \textbf{Sparsity} & \textbf{Highlight} \\
\hline
\multicolumn{7}{|c|}{\textbf{Segmentation}}\\
\hline
 Semantic3D \cite{hackel2017semantic3d}  & \multicolumn{1}{l|}{ASCII} & \begin{tabular}[c]{@{}c@{}}X, Y, Z, Intensity, \\ R, G, B\end{tabular} & \begin{tabular}[c]{@{}c@{}}4 billion \\ points\end{tabular} & 8 & Dense & \begin{tabular}[c]{@{}c@{}}training \& testing; \\ competing with the \\ most algorithms\end{tabular} \\ \hline
 Oakland \cite{munoz2009contextual} & ASCII & X, Y, Z, Class & \begin{tabular}[c]{@{}c@{}}1.6 million \\ points\end{tabular} & 5 & Sparse & \begin{tabular}[c]{@{}c@{}}training to tune \\ model architecture\end{tabular} \\ \hline
 iQmulus \cite{vallet2015terramobilita} & PLY & \begin{tabular}[c]{@{}c@{}}X, Y, Z, Intensity, \\ GPS time, Scan origin,\\ \# echoes,  Object ID, \\ Class\end{tabular} & \begin{tabular}[c]{@{}c@{}}300.0 \\ million \\ points\end{tabular} & 22 & Moderate & training \& testing \\   \hline
 Paris-Lille-3D \cite{roynard2018classification} & PLY & \begin{tabular}[c]{@{}c@{}}X, Y, Z, \\ Intensity, \\  Class\end{tabular} & \begin{tabular}[c]{@{}c@{}}143.1 \\ million\\  points\end{tabular} & 50 & Moderate & \begin{tabular}[c]{@{}c@{}} \\training \& testing; \\ competing with limited\\ algorithms\end{tabular} \\ \hline

\multicolumn{7}{|c|}{\textbf{Localization/Detection}}\\
\hline
 \begin{tabular}[c]{@{}c@{}}KITTI Object  \\ Detection/\\ Bird's  Eye View  \\ Benchmark \cite{geiger2012we} \end{tabular} & - & \begin{tabular}[c]{@{}c@{}}3D bounding \\ boxes\end{tabular} & \begin{tabular}[c]{@{}c@{}}80,256 \\ objects\end{tabular} & 3 & Sparse & \begin{tabular}[c]{@{}c@{}}training \& testing; \\ competing with the \\most algorithms\end{tabular} \\ \hline

\multicolumn{7}{|c|}{\textbf{Classification/Recognition}}\\
\hline
 \begin{tabular}[c]{@{}c@{}}Sydney Urban \\ Objects Dataset \cite{de2013unsupervised}\end{tabular} & ASCII & \begin{tabular}[c]{@{}c@{}}Timestamp, Intensity,  \\ Lser id, X, Y, Z, \\ Azimuth,Range, id\end{tabular} & 588 objects & 14 & Sparse & \begin{tabular}[c]{@{}c@{}}training \& testing; \\ competing with limited\\ algorithms\end{tabular} \\ \hline
 \begin{tabular}[c]{@{}c@{}} ModelNet \cite{wu20153d} \end{tabular} & ASCII & \begin{tabular}[c]{@{}c@{}} X, Y, Z, \\ number of Vertices,\\ edges, faces \end{tabular} &  \begin{tabular}[c]{@{}c@{}} 12,311 objects (ModelNet40) \\ 4,899 objects (ModelNet10) \end{tabular} & \begin{tabular}[c]{@{}c@{}} 40 (ModelNet40) \\ 10 (ModelNet10) \end{tabular} & Dense & \begin{tabular}[c]{@{}c@{}}training \& testing; \\ competing with most\\ algorithms\end{tabular} \\ 
\hline
\end{tabular}
\label{t_dataset}
\end{table*}

As shown in Table \ref{t_dataset}, we classify those existing datasets related to our topic into three types: segmentation-based datasets, detection-based datasets, classification-based datasets. Besides, long-term autonomy dataset is also summarized.
\begin{itemize}

    \item Segmentation-based datasets
\end{itemize}

\textbf{Semantic3D} \cite{hackel2017semantic3d}. Semantic3D is the existing largest LiDAR dataset for outdoor scene segmentation tasks with more than 4 billion points and around 110,000$m^2$ covering area. This dataset is labeled with 8 classes and split into training and test sets with nearly equal size. These data are acquired by a static LiDAR with high measurement resolution and covered long measurement distance. The challenges for this dataset mainly stems from the massive point clouds, unevenly distributed point density, and severe occlusions. In order to fit the high computation algorithms, a reduced-8 dataset is introduced for training and testing, which share the same training data but fewer test data compared with Semantic3D.

\textbf{Oakland 3-D Point Cloud Dataset} \cite{munoz2009contextual}. This dataset is acquired in an early year compared with the above two datasets. A mobile platform equipped with LiDAR is used to scan the urban environment and generated around 1.3 million points, while 100,000 points are split into a validation set. The whole dataset is labeled with 5 classes such as wire, vegetation, ground, pole/tree-trunk, and facade. This dataset is small and thus suitable for lightweight networks. Besides, this dataset can be used to test and tune the network architectures without a lot of training time before final training on other datasets.

\textbf{IQmulus \& TerraMobilita Contest} \cite{vallet2015terramobilita}. This dataset is also acquired by a mobile LiDAR system in the urban environment in Paris. There are more than 300 million points in this dataset, which covered 10km street. The data is split into 10 separate zones and labeled with more than 20 fine classes. However, this dataset also has severe occlusion.

\textbf{Paris-Lille-3D} \cite{roynard2018classification}. Compared with Semantic3D \cite{hackel2017semantic3d}, Paris-Lille-3D contains fewer points (140 million points) and covering area (55,000$m^2$). The main difference of this dataset is that its data are acquired by a Mobile LiDAR system in two cities: Paris and Lille. Thus, the points in this dataset are sparse and comparatively low measurement resolution compared with Semantic3D \cite{hackel2017semantic3d}. But this dataset is more similar to the LiDAR data acquired by AVs. The whole dataset is fully annotated into 50 classes unequally distributed in three scenes:Lille1, Lille2, and Paris. For simplicity, these 50 classes are combined into 10 coarse classes for challenging.

\begin{table*}[!ht]
\centering
\caption{Evaluation metrics for 3D point cloud segmentation, detection/localization, and classification}
\begin{tabular}{|l|l|l|}
\hline
\textbf{Metric} &  \multicolumn{1}{c|}{\textbf{Equation}} &  \multicolumn{1}{c|}{\textbf{Description}} \\ \hline
 
{$IoU$} & \scalebox{1.0}{${IoU_{i}}$}= \scalebox{1.0}{$\frac{c_{i i}}{c_{i i}+\sum_{j \neq i} c_{i j}+\sum_{k \neq i} c_{k i}}$} & \begin{tabular}[l]{@{}l@{}} Intersection over Union, where $c_{ij}$ is the number of points from ground-truth\\ class $i$ predicted as class $j$ \cite{everingham2015pascal} \end{tabular}\\ \hline

{$\overline{{IoU}}$} & \scalebox{1.0}{$ \overline{{IoU}}$}=\scalebox{1.0}{$\frac{\sum_{i=1}^{N} I o U_{i}}{N}$}  & Mean IoU, where N is the number of classes \\ \hline

{$\mathrm{{OA}}$} & \scalebox{1.0}{$\mathrm{{OA}}$}= \scalebox{1.0}{$ \frac{\sum_{i=1}^{N} c_{i i}}{\sum_{j=1}^{N} \sum_{k=1}^{N} c_{j k}}$ }  & Overall accuracy\\ \hline

{${Precision}$} & \scalebox{1.0}{${Precision}$} = \scalebox{1.0}{$\frac{T P}{T P+F P}$}  &  \begin{tabular}[l]{@{}l@{}} The ratio of correctly detected objects in the whole detection results, where \\ $TP,TN,FP$,and $FN$ are the number of true positives, true negatives, false \\ positives,and false negatives, respectively \cite{yan2017detection} \end{tabular} \\ \hline

{${Recall}$} & \scalebox{1.0}{${Recall}$}=\scalebox{1.0}{$\frac{TP}{(TP+FN)}$}  & \begin{tabular}[l]{@{}l@{}} The ratio of correctly detected objects in the ground truth  \end{tabular} \\ \hline

{${F_1}$} & \scalebox{1.0}{${F_1}$}=\scalebox{1.0}{$\frac{2TP}{(2TP+FP+FN)} $ } &  \begin{tabular}[l]{@{}l@{}}  The balance between precision and recall  \end{tabular} \\ \hline

{${MCC}$} & \scalebox{1.0}{${MCC}$}=\scalebox{1.0}{$\frac{(TP\times TN-FP\times FN)}{\sqrt{((TP+FP)(TP+FN)(TN+FP)(TN+FN))}}$ } & \begin{tabular}[l]{@{}l@{}} The combined  ratio of detected and undetected objects as well as non-objects \end{tabular}\\ \hline

{$\mathrm{{AP}}$} & \scalebox{1.0}{$\mathrm{{AP}}$}=\scalebox{1.0}{$\frac{1}{11} \sum_{r \in\{0,1, \ldots, 1\}} \max _{\tilde{r} : \tilde{r} \geq r} p(\tilde{r}) $ } &  Average Precision, where $r$ represents the recall, $p_(r)$ represents the precision  \\ \hline

{${AOS}$} & \scalebox{1.0}{${AOS}$}=\scalebox{1.0}{$\frac{1}{11} \sum_{r \in\{0,0.1, \ldots 1\}} \max _{\overline{r} ; \vec{r} \geq r} s(\tilde{r})$}, & Average Orientation Similarity\\ \hline

{$s(r)$} & \scalebox{1.0}{$ s(r)$} =\scalebox{1.0}{$s(r)=\frac{1}{|\mathcal{D}(r)|} \sum_{i \in \mathcal{D}(r)} \frac{1+\cos \Delta_{\theta}^{(i)}}{2} \delta_{i}$} &  \begin{tabular}[l]{@{}l@{}} Orientation similarity, where $\mathcal{D}(r)$ represents the whole object detection\\ at recall rate $r$ and $\Delta_{\theta}^{(i)}$ is the angle difference between predicted and \\ ground truth orientation of detection $i$, $\delta_{i}$ is the penalty value when \\multiple detection tasks describe one object \end{tabular} \\ \hline
\end{tabular}
\label{t_evaluation}
\end{table*}

\begin{itemize}
    \item Detection-based datasets
\end{itemize}

\textbf{KITTI Object Detection/Bird’s Eye View Benchmark} \cite{geiger2012we}. Different from the above LiDAR datasets which are specific for segmentation task, KITTI dataset is acquired from an autonomous driving platform and records six hours driving using digital cameras, LiDAR, GPS/IMU inertial navigation system. Thus, apart from the LiDAR data, the corresponding imagery data are also provided. Both the Object Detection and Bird’s Eye View Benchmark contains 7481 training images and 7518 test images as well as the corresponding point clouds. Due to the moving scanning mode, the LiDAR data in this benchmark is highly sparse. Thus, only three objects are labeled with bounding box: cars, pedestrians, and cyclists.

\begin{itemize}
    \item Classification-based datasets
\end{itemize}

\textbf{Sydney Urban Objects Dataset} \cite{de2013unsupervised}. This dataset contains a set of general urban road objects scanned with a LiDAR in the CBD of Sydney, Australia. There are 588 labeled objects and classified in 14 categories, such as vehicles, pedestrians, signs, and trees. The whole dataset is split into four folds for training and testing. Similar to other LiDAR datasets, the collected objects in this dataset are sparse with incomplete shape. Although it is small and not ideal for the classification task, it the most commonly used benchmark due to the limitation of the tedious labeling process.

\textbf{ModelNet} \cite{wu20153d}. This dataset is the existing largest 3D benchmark for 3D object recognition. Different from Sydney Urban Objects Dataset \cite{de2013unsupervised}, which contains road objects collected by LiDAR sensors, this dataset is composed of general objects in CAD models with evenly distributed point density and complete shape. There are approximately 130K labeled models in a total of 660 categories (e.g., car, chair, clock). The most commonly used benchmarks are ModelNet40 that contains 40 general objects and ModelNet10 with 10 general objects. The milestone 3D deep architectures are commonly trained and tested on these two datasets due to the affordable computation burden and time. 

\textbf{Long-Term Autonomy}: To address challenges of long-term autonomy, a novel dataset for autonomous driving has been presented by Maddern et al. \cite{maddern20171}. They collected images, LiDAR, and GPS data while traversing 1,000 km in central Oxford in the UK for one year. This allowed them to capture different scene appearances under various illumination, weather, and season with dynamic objects and constructions. Such long-term datasets allow for in-depth investigation of problems that detain the realization of autonomous vehicles such as localization at different times of the year.

\subsection{Evaluation Metrics}
To evaluate those proposed methods performance, several metrics, as summarized in Table \ref{t_evaluation}, are proposed for those tasks: segmentation, detection, and classification. The detail of these metrics is given as follows.

For the segmentation task, the most commonly used evaluation metrics are the Intersection over Union (IoU) metric, $\overline{IoU}$, and overall accuracy (OA) \cite{everingham2015pascal}. IoU defines the quantify the percent overlap between the target mask and the prediction output \cite{hackel2017semantic3d}. 

For detection and classification tasks, the results are commonly analyzed region-wise. Precision, recall, $F_1$-score and Matthews correlation coefficient (MCC) \cite{zhang2018large} are commonly used to evaluate the performance. The precision represents the ratio of correctly detected objects in the whole detection results, while the recall means the percentage of the correctly detected objects in the ground truth, the $F_1$-score conveys the balance between the precision and the recall, the MCC is the combined ratio of detected and undetected objects and non-objects. 

For 3D object localization and detection task, the most frequently used metrics are: Average Precision ($AP_{3D}$) \cite{chen20183d}, and Average Orientation Similarity (AOS) \cite{arnold2019survey}. The average precision is used to evaluate the localization and detection performance by calculating the averaged valid bounding box overlaps, which exceed predefined values. For orientation estimation, the orientation similarities with different threshold-ed valid bounding box overlaps are averaged to report the performance.

\section{General 3D Deep Learning Frameworks}
In this section, we review the milestone DL frameworks on 3D data. These frameworks are pioneers in solving the problems defined in section II. Besides, their stable and efficient performance makes them suitable for use as the backbone framework in detection, segmentation and classification tasks. Although 3D data acquired by LiDAR is often in the form of point clouds, how to represent point cloud and what DL models to use for detection, segmentation and classifications remains an open problem \cite{qi2018frustum}. Most existing 3D DL models process point clouds mainly in form of voxel grids \cite{wu20153d,maturana2015voxnet,wu2016learning,riegler2017octnet}, point clouds  \cite{qi2017pointnet,qi2017pointnet++,klokov2017escape,li2018pointcnn}, graphs \cite{yi2017syncspeccnn,simonovsky2017dynamic,wang2018dynamic,velivckovic2017graph} and  2D images \cite{su2015multi,qi2016volumetric,dai20183dmv,kanezaki2018rotationnet}. In this section, we analyze the frameworks, attributes and problems of these models in detail.

\begin{figure}[!t]
\centering
\includegraphics[scale=0.35]{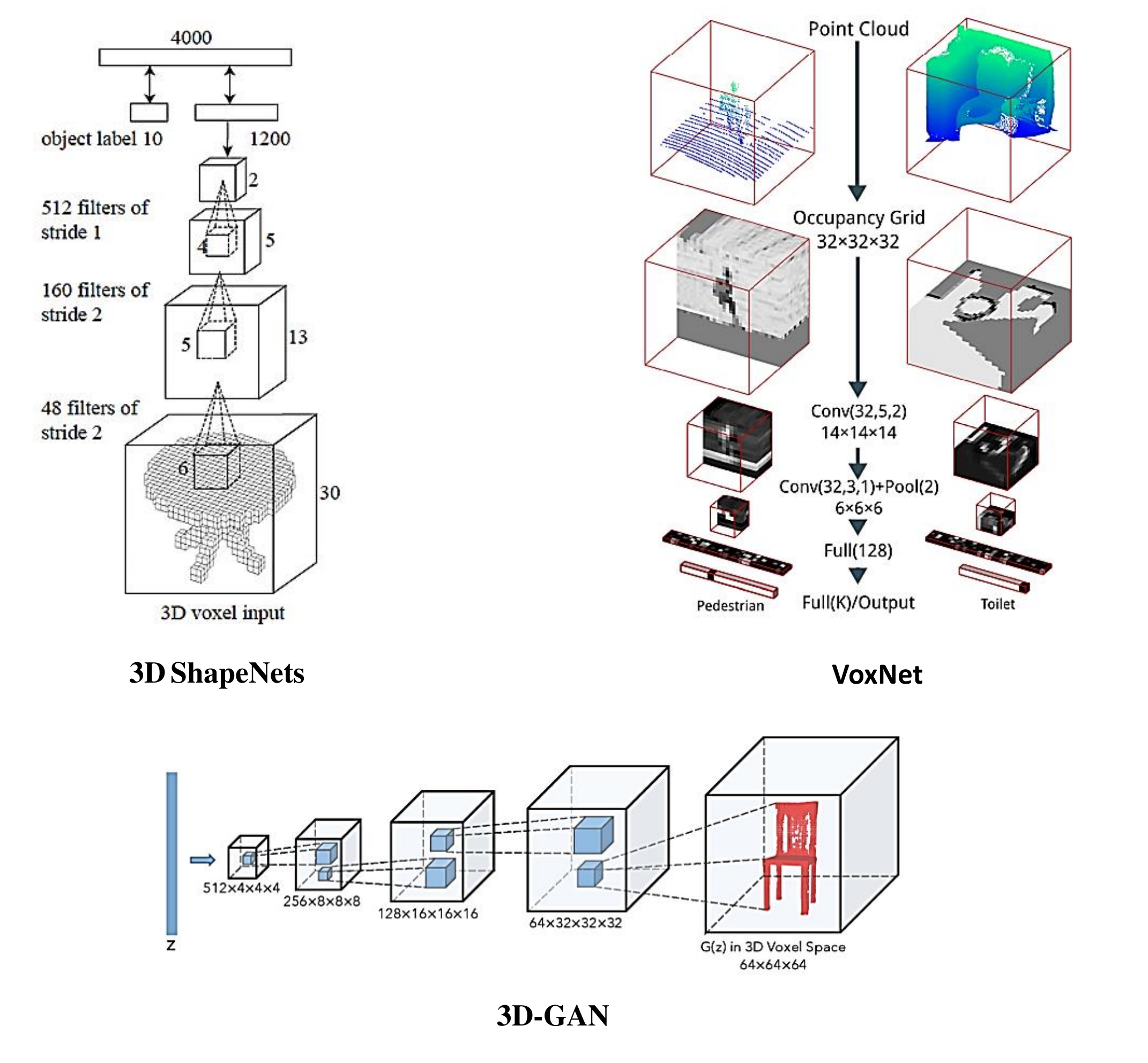}
\caption{Deep architectures of 3D ShapeNet \cite{wu20153d}, VoxNet \cite{maturana2015voxnet}, 3D-GAN \cite{wu2016learning}.}
\label{fig_shape_vox_gan_oct}
\end{figure}

\subsection{Voxel-based models}
Conventionally, CNNs are mainly applied to data with regular structures, such as the 2D pixel array \cite{long2015fully}. Thus, in order to apply CNNs to unordered 3D point cloud data, such data are  divided into regular grids with a certain size to describe the distribution of data in 3D space. Typically, the size of the grid is related to the resolution of data \cite{vosselman2004recognising}. The advantage of voxel-based representation is that it can encode the 3D shape and viewpoint information by classifying the occupied voxels into several types such as visible, occluded, or self-occluded. Besides, 3D convolution (Conv) and pooling operations can be directly applied in voxel grids \cite{riegler2017octnet}. 

\textbf{3D ShapeNet} \cite{wu20153d}, proposed by Wu et al. and shown in Fig.\ref{fig_shape_vox_gan_oct}, is the pioneer in exploiting 3D volumetric data using a convolutional deep belief network. The probability distribution of binary variables is used to represent the geometric shape of a 3D voxel grid. Then these distributions are input to the network which is mainly composed of three Conv layers. This network is initially pre-trained in a layer-wise fashion and then trained with a generative fine-tuning procedure. The input and Conv layers are modeled based on the Contrastive Divergence, where the output layer was trained based on the Fast-Persistent Contrastive Divergence. After training, the input test data is output with a single depth map and then transformed to represent the voxel grid. ShapeNet has notable results in low-resolution voxels. However, the computation cost increases cubically with the increment of input data size or resolution, which limit the model’s performance in large-scale or dense point clouds data. Besides, multi-scale and multi-view information from the data is not fully exploited, which hinder the output performance.

\textbf{VoxNet} \cite{maturana2015voxnet} is proposed by Maturana et al. to conduct 3D object recognition using 3D convolution filters based on volumetric data representation, as shown in Fig.\ref{fig_shape_vox_gan_oct}. Occupancy grids represented by a 3D lattice of random variables are employed to show the state of the environment. Then a probabilistic estimate is used to estimate the occupancy of these grids which is maintained as the prior knowledge. Three different occupancy grid models, such as binary occupancy grid, density grid, and hit grid are experimented to select the best model. This network framework is mainly composed of Conv, pooling layer, and fully connected (FC) layers. Both ShapeNet \cite{wu20153d} and VoxNet employ rotation augmentation for training. Compared with ShapeNet \cite{wu20153d}, VoxNet has a smaller architecture that has less than 1 million parameters. However, not all occupancy grids contain useful information but only increase the computation cost.

\textbf{3D-GAN} \cite{wu2016learning} combines the merits of both general-adversarial network (GAN) \cite{goodfellow2014generative} and volumetric convolutional networks \cite{maturana2015voxnet} to learn the features of 3D objects. This network is composed of a generator and a discriminator as shown in Fig.\ref{fig_shape_vox_gan_oct}. The adversarial discriminator is conducted to classify objects into synthesized and real categories due to the generative-adversarial criterion has the advantage in capturing the structural variation between two 3D objects. And the employment of generative-adversarial loss is helpful to avoid possible criterion-dependent over-fitting. The generator attempts to confuse the discriminator. Both generator and discriminator consist of five volumetric fully Conv layers. This network provides a powerful 3D shape descriptor with unsupervised training in 3D object recognition. But the density of data affects the performance of adversarial discriminator for finest feature capturing. Consequently, this adaptive method is suitable for evenly distributed point cloud data.

In conclusion, there are some limitations of this general volumetric 3D data representation:
\begin{itemize}
\item Firstly, not all voxel representations are useful because they contain occupied and non-occupied parts of the scanning environment. Thus, the high demand for computer storage is actually unnecessary within this ineffective data representation \cite{riegler2017octnet}.
\item Secondly, the size of the grid is hard to set, which affects the scale of input data and may disrupt the spatial relationship between points. 
\item Thirdly, computational and memory requirements grow cubically with the resolution \cite{riegler2017octnet}. Thus, existing voxel-based models are maintained at low 3D resolutions, and the most commonly used size is $30^3$ for each grid.\cite{riegler2017octnet}.
\end{itemize}

\begin{figure}[!t]
\centering
\includegraphics[width=3.5in]{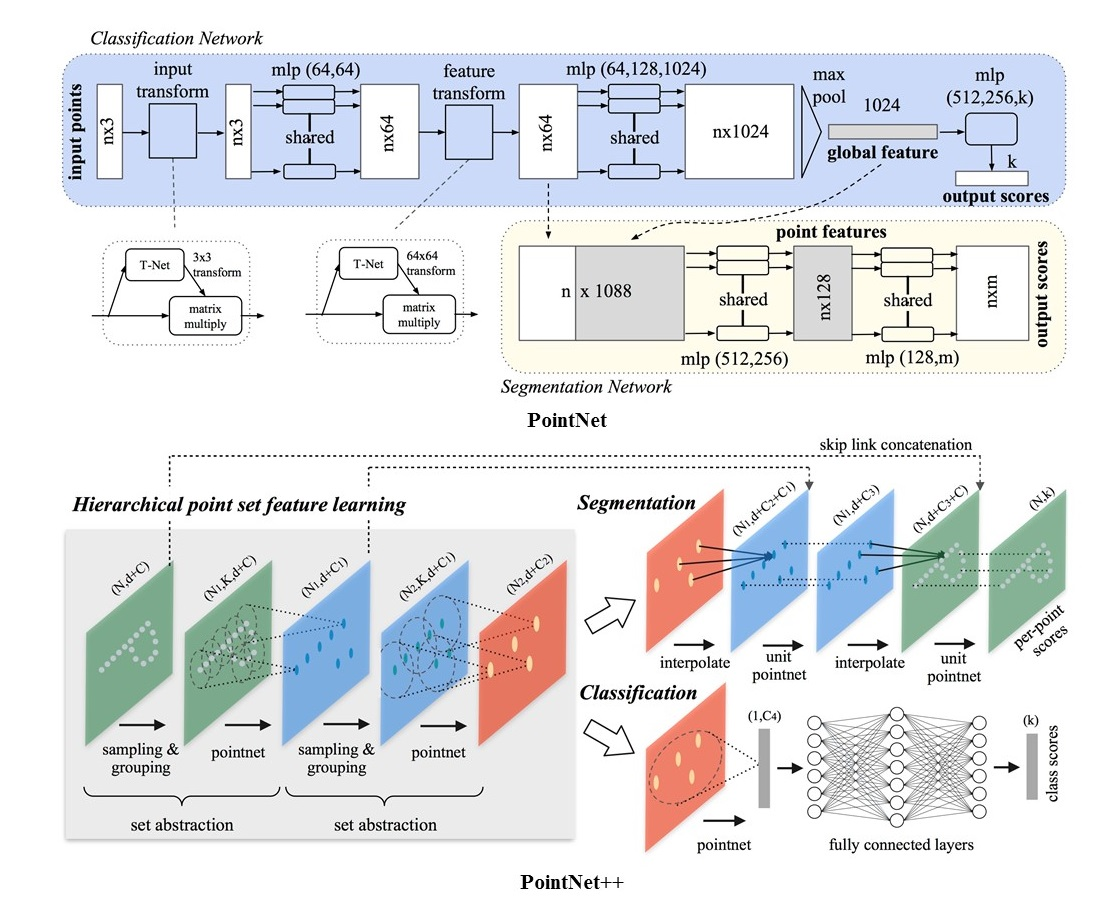}
\caption{PointNet \cite{qi2017pointnet} and PointNet++ \cite{qi2017pointnet++} architectures.}
\label{fig_pointnet&2}
\end{figure}

A more advanced voxel-based data representation is the octree-based grids \cite{riegler2017octnet,tatarchenko2017octree}, which use adaptive size to divides the 3D point cloud into cubes. It is a hierarchical data structure that recursively decomposes the root voxels into multiple leaf voxels. 

\textbf{OctNet} \cite{riegler2017octnet} is proposed by Riegler et al., which exploits the sparsity of the input data. Motivated by the observation that the object boundaries have the highest probability in producing the maximum responses across all feature maps generated by the network at different layers, they partitioned the 3D space hierarchically into a set of unbalanced octrees \cite{miller2011real} based on the density of the input data. Specifically, the octree nodes that have point clouds are split recursively in its domain, ending at the finest resolution of the tree. Thus, the size of leaf nodes varies. For each leaf node, those features that activate their comprised voxel is pooled and stored. Then the convolution filters are conducted in these trees. In \cite{tatarchenko2017octree}, the deep model is constructed by learning the structure of the octree and the represented occupancy value for each grid. This octree-based data representation largely reduces the computation and memory resources for DL architectures, which achieves better performance in high-resolution 3D data compared with voxel-based models. However, the disadvantage of octree data is similar to voxels, both of them fail to exploit the geometry feature of 3D objects, especially the intrinsic characteristics of patterns and surfaces \cite{ahmed2018deep}.

\subsection{Point clouds based models}

Different from volumetric 3D data representation, point cloud data can preserve the 3D geospatial information and internal local structure. Besides, the voxel-based models that scan the space with fixed strides are constrained by the local receptive fields. But for point clouds, the input data and the metric decide the range of receptive fields, which has high efficiency and accuracy.

\textbf{PointNet} \cite{qi2017pointnet}, as a pioneer in consuming 3D point clouds directly for deep models, learns the spatial feature of each point independently via MLP layers and then accumulates their features by max-pooling. The point cloud data are input directly to the PointNet, which predicts per-point label or per-object label, its framework is illustrated in Fig.\ref{fig_pointnet&2}. In PointNet, spatial transform network and a symmetric function are designed to improve the invariance to permutation. The spatial feature of each input point was learned through the networks. Then, the learned features are assembled across the whole region of point clouds. The outstanding performance of PointNet has achieved in 3D objects classification and segmentation tasks. However, the individual point features are grouped and pooled by max-pooling, which fails to preserve the local structure. As a result, PointNet is not robust to fine-grained patterns and complex scenes.

\textbf{PointNet++} was proposed later by Qi et al. \cite{qi2017pointnet++}, which compensate the local feature extraction problems in PointNet. Within the raw unordered point clouds as input, these points are initially divided into overlapping local regions using the Euclidean distance metric. These partitions are defined as a neighborhood ball in this metric space and labeled with the centroid location and scale. In order to sample the points evenly over the whole point set, the farthest point sampling (FPS) algorithm is applied. Local features are extracted from the small neighborhoods around the selected points using K-nearest-neighbor (KNN) or query-ball searching methods. These neighborhoods are gathered into larger clusters and leveraged to extract high-level features via PointNet \cite{qi2017pointnet} network. The sampling and grouping module are repeated until the local and global features of the whole points are learned, as shown in Fig.\ref{fig_pointnet&2}. This network, which outperforms the PointNet \cite{qi2017pointnet} network in classification and segmentation tasks, extracts the local feature for points in different scales. However, features from the local neighborhood points in different sampling layers are learned in an isolated fashion. Besides, max-pooling operation based on PointNet \cite{qi2017pointnet} for high-level feature extraction in PointNet++ fails to preserve the spatial information between the local neighborhood points. 

\begin{figure}[!t]
\centering
\includegraphics[width=2.5in]{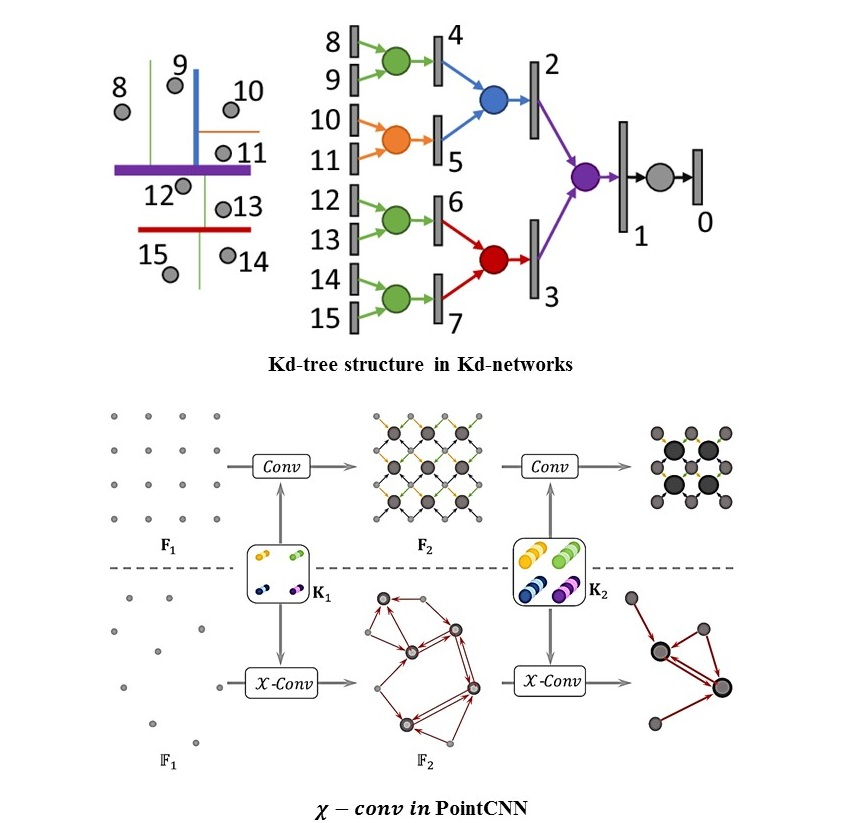}
\caption{Kd-tree structure in Kd-networks \cite{klokov2017escape} and $\chi$-Conv in PointCNN \cite{li2018pointcnn}.}
\label{fig_pointcnn_kd}
\end{figure}

\textbf{Kd-networks} \cite{klokov2017escape} uses the kd-tree to create the order of the input points, which is different from PointNet \cite{qi2017pointnet} and PointNet++ \cite{qi2017pointnet++} as both of them use the symmetric function to solve the permutation problem. Klokov et al. used the maximum range of point coordinates along the coordinate axis to recursively split the certain size point clouds $N=2^D$ into subsets with a top-down fashion to construct a kd-tree. As shown in Fig.\ref{fig_pointcnn_kd}, this kd-tree is ending with a fixed depth. Within this balanced tree structure, vectorial representations in each node, which represents a subdivision along certain axis, is computed using kd-networks. These representations are then exploited to train a linear classifier. This network has better performance than PointNet \cite{qi2017pointnet} and PointNet++ \cite{qi2017pointnet++} in small objects classification. However, it is not robust to rotations and noise, since these variations can lead to the change of tree structure. Besides, it lacks the overlapped receptive field which reduces the spatial-correlation between leaf nodes.

\textbf{PointCNN}, proposed by Li et al. \cite{li2018pointcnn}, solves the input points permutation and transformation problems based on an $\chi$-Conv operation, as shown in Fig.\ref{fig_pointcnn_kd}. They proposed the $\chi$-transformation which is learned from the input points by weighting the input point features and permutating the points into a latent and potentially canonical order. Then the traditional convolution operators are applied in the learned $\chi$-transformation features. These spatially-local correlation features in each local range are aggregated to construct a hierarchical CNN network architecture. However, this model still has not exploited the correlations of different geometric features and their discriminate information toward results, which limits the performance. 

Point cloud based deep models are mostly focused on solving permutation problems. Although they treat points independently at local scales to maintain permutation invariance. This independence, however, neglects the geometric relationships among points and their neighbors, presenting a fundamental limitation that leads to local features' missing.

\subsection{Graph-based models}
Graphs are a type of non-Euclidean data structure that can be used to represent point cloud data. Their node corresponds to each input point and the edges represent the relationship between each point neighbors. Graph neural networks propagate the node states until equilibrium in an iterative manner \cite{velivckovic2017graph}. With the advancement of CNNs, there is an increment graph convolutional networks applied to 3D data. Those graph CNNs define convolutions directly on the graph in the spectral and non-spectral (spatial) domain, operating on groups of spatially close neighbors \cite{wang2018local}. The advantage of graph-based models is that the geometric relationships among points and their neighbors are exploited. Thus, more spatially-local correlation features are extracted from the grouped edge relationships on each node. But there are two challenges for constructing graph-based deep models:
\begin{itemize}
    \item Firstly, defining an operator that is suitable for dynamically sized neighborhoods and maintaining the weight sharing scheme of CNNs \cite{velivckovic2017graph}.
    \item Secondly, exploiting the spatial and geometric relationships among each node's neighbors. 
\end{itemize}

\textbf{SyncSpecCNN} \cite{yi2017syncspeccnn} exploited the spectral eigen-decomposition of the graph Laplacian to generate a convolution filter applied in point clouds. Yi et al. constructed SyncSpecCNN based on that two considerations: the first is the coefficients sharing and multi-scale graph analyzing; the second is information sharing across related but different graphs. They solved these two problems by constructing the convolution operation in the spectral domain: the signal of point sets in the Euclidean domain is defined by the metrics on the graph nodes, and the convolution operation in the Euclidean domain is related to the scaling signals based on eigenvalues. Actually, such operation is linear and only applicable to the graph weights generated from eigenvectors of the graph Laplacian. Despite SyncSpecCNN achieved excellent performance in 3D shape part segmentation, it has several limitations:
\begin{itemize}
    \item Basis-dependent. The learned spectral filter’s coefficients are not suitable for another domain with a different basis.
    \item Computationally expensive. The spectral filtering is calculated based on the whole input data, which requires high computation capability.
    \item Missing local edge features. The local graph neighborhood contains useful and distinctive local structural information, which is not exploited.
\end{itemize}

\textbf{Edge-conditioned convolution} (ECC) \cite{simonovsky2017dynamic} considers the edge information in constructing the convolution filters based on the graph signal in the spatial domain. The edge labels in a vertex neighborhood are conditioned to generate the Conv filter weights. Besides, in order to solve the basis-dependent problem, they dynamic generalized the convolution operator for arbitrary graphs with varying size and connectivity. The whole network follows the common structure of feedforward network with interlaced convolutions and pooling followed by global pooling and FC layers. Thus, features from local neighborhoods are extracted continually from these stacked layers, which increase the receptive field. Although the edge labels are fixed for a specific graph, the learned interpretation networks may vary in different layers. ECC learns the dynamic pattern of local neighborhoods, which is scalable and effective. However, the computation cost remains high, and it is not applicable for large-scale graphs with continuous edge labels. 

\begin{figure}[!t]
\centering
\includegraphics[width=2.5in]{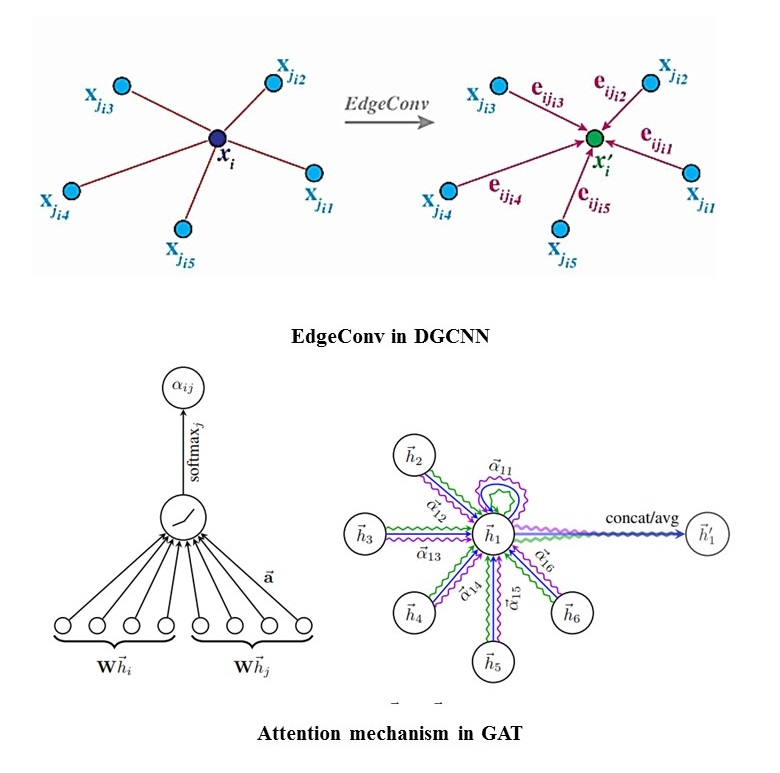}
\caption{EdgeConv in DGCNN \cite{wang2018dynamic} and attention mechanism in GAT \cite{velivckovic2017graph}.}
\label{fig_dgcnn_gat}
\end{figure}

\textbf{DGCNN} \cite{wang2018dynamic} also constructed a local neighborhood graph to extract the local geometric features and applied Conv-like operations, named EdgeConv which is shown in Fig.\ref{fig_dgcnn_gat}, on the edges connecting neighboring pairs of each point. Different from ECC \cite{simonovsky2017dynamic}, EdgeConv dynamically updates the given fixed graph with Conv-like operations for each layer output. Thus, DGCNN can learn how to extract local geometric structures and group point clouds. This model takes $n$ points as input, and then find the K neighborhoods of each point to calculate the edge feature between the point and its K neighborhoods in each EdgeConv layer. Similar to PointNet[34] architecture, the features convolved in the last EdgeConv layer are aggregated globally to construct a global feature, while all the EdgeConv outputs are treated as local features. Local and global features are concatenated to generate results’ score. This model extracts distinctive edge features from point neighborhoods, which can be applied in different point clouds related tasks. However, the fixed size of edge features limits the performance of the model when facing different scales and resolution point clouds.

ECC \cite{simonovsky2017dynamic} and DGCNN \cite{wang2018dynamic} propose general convolutions on graph nodes and their edge information, which is isotropy about input features. However, not all the input features contribute equally to its nodes. Thus, attention mechanisms are introduced to deal with variable sized inputs and focus on the most relevant parts of the nodes' neighbors to make decisions \cite{velivckovic2017graph}. 

\textbf{Graph Attention Networks} (GAT) \cite{velivckovic2017graph}. The core insight behind GAT is to calculate the hidden representations of each node in the graph, by assigning different attentional weights to different neighbors, following a self-attention strategy. Within a set of node features as input, a shared linear transformation, parametrized by a weight matrix is applied to each node. Then a self-attention, a shared attentional mechanism which is shown in Fig.\ref{fig_dgcnn_gat}, is applied on the nodes to computes attention coefficients. These coefficients indicate the importance of corresponding nodes' neighbor features, respectively, and are further normalized to make them comparable across different nodes. These local features are combined according to the attentional weights to form the output features for each node. In order to improve the stability of the self-attention mechanism, multi-head attention is employed to conduct k independent attention schemes, which are then concatenated together to form the final output features for each node. This attention architecture is efficient and can extract fine-grained representations for each graph node by assigning different weights to the neighbors. However, local spatial relationship between neighbors are not considered in calculating the attentional weights. To further improve its performance, Wang et al. \cite{wang2019graph} proposed graph attention convolution (GAC) to generate attentional weights by considering different neighboring points and feature channels.

\begin{table*}[!ht]
\caption{Summarizing of milestone DL architectures based on four point cloud data representations}
\centering
\begin{tabular}{|l|c|l|l|c|c|}
\hline 
 \textbf{Model} & \textbf{\begin{tabular}[|c|]{@{}c@{}}Input  \\ Size\end{tabular}} & \multicolumn{1}{c|}{\textbf{Hightlights}} & \multicolumn{1}{c|}{\textbf{Disadvanatges}} &\textbf{\begin{tabular}[c|]{@{}c@{}}Model \\ size(MB)\end{tabular}} &\textbf{\begin{tabular}[c|]{@{}c@{}}Acc \\ (\%)\end{tabular}} \\ \hline
\multicolumn{6}{|c|}{\textbf{Voxel}}\\
\hline
 \begin{tabular}[c]{@{}l@{}}3dShapeNet\\ \cite{wu20153d}\end{tabular}  & voxels & \begin{tabular}[c]{@{}l@{}}Pioneer in exploiting 3D volumetric data;\\ Permutation and orientation invariance. \end{tabular} & \begin{tabular}[c]{@{}l@{}} Computation and memory requirement grows \\cubically; Use one view in a fixed voxel size.\end{tabular}& ~12 & 84.7 \\ \hline
  \begin{tabular}[c]{@{}l@{}}VoxNet\\ \cite{maturana2015voxnet}\end{tabular}  & voxels & \begin{tabular}[c]{@{}l@{}}Occupancy grids are employed to represent\\ the distribution of the scene as a 3D lattice \\ of random variables for each grid; Permutation \\and orientation invariance; Improved efficiency. \end{tabular} &\begin{tabular}[c]{@{}l@{}} Not all occupancies are useful.\end{tabular}& ~1.0 & 85.9 \\ \hline
  \begin{tabular}[c]{@{}l@{}}3D-GAN\\ \cite{wu2016learning}\end{tabular}  & voxels & \begin{tabular}[c]{@{}l@{}}Combines the adversarial modeling and volumetric \\ convolutional networks to learn features; \\ Permutation and orientation invariance;\\ Rigid transformation invariance.\end{tabular} &\begin{tabular}[c]{@{}l@{}} Not invariance to data density variation\end{tabular}& ~7 & 83.3 \\ \hline
 \begin{tabular}[c]{@{}l@{}}OctNet\\ \cite{riegler2017octnet}\end{tabular}  & \begin{tabular}[c]{@{}c@{}}hybrid \\ grid \\ octree\end{tabular} & \begin{tabular}[c]{@{}l@{}}Hierarchically divide the data into a series of \\ unbalanced octrees according to data density;\\ Permutation and orientation invariance; Efficient.\\\end{tabular} & \begin{tabular}[c]{@{}l@{}}Fail to preserve the geometry relationship \\ among points.\end{tabular} & 0.4 & 86.5 \\ \hline
\multicolumn{6}{|c|}{\textbf{Point Clouds}}\\
\hline
  \begin{tabular}[c]{@{}l@{}}PointNet\\ \cite{qi2017pointnet}\end{tabular}  & \begin{tabular}[c]{@{}c@{}}1024 \\ points\end{tabular} & \begin{tabular}[c]{@{}l@{}}Pioneer in applying DL using 3D \\ point clouds and solving the \\permutation problem via maxpooling. \end{tabular} & \begin{tabular}[c]{@{}l@{}}Not capture local structure induced by \\ the metric; Hard to generalize to unseen \\ point configurations.\end{tabular} & 40~ & 89.2 \\ \hline
\begin{tabular}[c]{@{}l@{}}PointNet++\\ \cite{qi2017pointnet++}\end{tabular}  & \begin{tabular}[c]{@{}c@{}}5000 \\ points \\ +normal\end{tabular} & \begin{tabular}[c]{@{}l@{}}Hierarchically learn multi-scale local geometric \\ features and aggregate them for inference; \\Permutation and rigid transformation invariance \\and efficient.\end{tabular} & \begin{tabular}[c]{@{}l@{}} Local spatial relationship among point \\ neighborhoods is not exploited. \end{tabular} & 12 & 90.7 \\ \hline
 \begin{tabular}[c]{@{}l@{}}Kd-networks \\ \cite{klokov2017escape} \end{tabular} & \begin{tabular}[c]{@{}c@{}}1024 \\ points\end{tabular} & \begin{tabular}[c]{@{}l@{}}Use the kd-tree to create the order of  the input \\ points and hierarchically extract features \\from the leaves to root; Permutation and rigid \\transformation invariance.\end{tabular} & \begin{tabular}[c]{@{}l@{}}Non-invariance to rotations and noises;\\ Computation grows linearly with increasing \\ resolution; Low spatial-correlation between \\ leaf nodes.\end{tabular} & ~120 & 91.8 \\ \hline
  \begin{tabular}[c]{@{}l@{}}PointCNN\\ \cite{li2018pointcnn} \end{tabular}  & \begin{tabular}[c]{@{}c@{}}1024 \\ points\end{tabular} & \begin{tabular}[c]{@{}l@{}}Propose X-Conv operator that permutes \\ and weights input points and features;\\ Permutation and rigid transformation invariance. \end{tabular} & \begin{tabular}[c]{@{}l@{}}Not exploit the correlations of different \\ geometric features and their discriminative \\ information toward final results.\end{tabular} & 4.5 & 92.2 \\ \hline
\multicolumn{6}{|c|}{\textbf{Graph}}\\
\hline
 \begin{tabular}[c]{@{}c@{}}Spectral\\ -CNN \cite{yi2017syncspeccnn}\end{tabular} & graphs & \begin{tabular}[c]{@{}l@{}}Exploit the spectral eigen-decomposition of the \\graph Laplacian to generate a Conv-like operator.\end{tabular} & \begin{tabular}[c]{@{}l@{}}Basis-dependent; Computationally expensive; \\ Missing local edge features.\end{tabular} & 0.8 & - \\ \hline
 \begin{tabular}[c]{@{}l@{}}ECC\\ \cite{simonovsky2017dynamic}  \end{tabular} & graphs & \begin{tabular}[c]{@{}l@{}} The edge labels in a vertex neighborhood \\ are conditioned to generate the Conv \\ filter weights; Permutation invariance.\end{tabular} & \begin{tabular}[c]{@{}l@{}}High computation cost; Not suitable for \\ large-scale graphs with continuous labels;\\ Isotropic about input features.\end{tabular} & -& 87.4 \\ \hline
  \begin{tabular}[c]{@{}l@{}}DGCNN\\ \cite{wang2018dynamic} \end{tabular}  & graphs & \begin{tabular}[c]{@{}l@{}}Extract edge features and dynamically update the\\ graph for each layer; Permutation invariance.\end{tabular} & \begin{tabular}[c]{@{}l@{}}Fixed size edge features are not invariance \\ to points with different resolution \\ and scale; Isotropic about input features.\end{tabular} & ~21 &92.2 \\ \hline
 \begin{tabular}[c]{@{}l@{}}GAT\\ \cite{velivckovic2017graph} \end{tabular}  & graphs & \begin{tabular}[c]{@{}l@{}}Compute the hidden representations of each node's \\neighbors, following a self-attention strategy;\\Permutation and invariance, improved accuracy.\end{tabular} & \begin{tabular}[c]{@{}l@{}} Apply the attention mechanism only to input \\ points not to their local features.\end{tabular} &- &- \\ \hline
\multicolumn{6}{|c|}{\textbf{2D View}}\\
\hline
 \begin{tabular}[c]{@{}l@{}}MVCNN\\ \cite{su2015multi} \end{tabular} & 12 views & \begin{tabular}[c]{@{}l@{}} Pioneer in applying CNN to each view and then \\ aggregate the features by a view pooling procedure;\\Permutation and orientation invariance; Efficient.\end{tabular} & \begin{tabular}[c]{@{}l@{}}Multi-resolution features are not \\ considered.\end{tabular} & 99 &90.1 \\ \hline
 \begin{tabular}[l]{@{}l@{}}MVCNN\\ -MultiRes \\\cite{qi2016volumetric}\end{tabular} & 20 views & \begin{tabular}[c]{@{}l@{}}Propose multi-resolution 3D filtering to capture \\comprehensive information at multi-scales;\\ Permutation and orientation invariance; Efficient.\end{tabular} & Geometric information are not exploited. & ~16.6  &91.4 \\ \hline
\begin{tabular}[c]{@{}l@{}}3DMV\\ \cite{dai20183dmv} \end{tabular} & 20 views & \begin{tabular}[c]{@{}l@{}}Extract RGB and geometric features and aggregate\\ them via a joint 2D-3D network; Permutation and \\orientation invariance; Efficient.\end{tabular} &\begin{tabular}[c]{@{}l@{}}2D occlusion and background clutter\\ affects the 3D network performance. \end{tabular} & - &- \\ \hline
\begin{tabular}[c]{@{}l@{}}RotationNet\\ \cite{kanezaki2018rotationnet} \end{tabular}  & 12 views & \begin{tabular}[c]{@{}l@{}}Treat the viewpoints of the observed  training images \\as latent variables; Permutation and orientation \\invariance; High accuracy. \end{tabular} & Not suitable for per-point processing tasks. & 59 & \textbf{97.37} \\ \hline 
\end{tabular}
\label{t2}
\end{table*}

\subsection{View-based models}
The last type of MLS data representation is 2D views obtained from 3D point clouds from different directions. With the projected 2D views, traditional well-established convolutional neural networks (CNN) and pre-trained networks on image datasets, such as AlexNet \cite{krizhevsky2012imagenet}, VGG \cite{simonyan2014very}, GoogLeNet \cite{szegedy2015going}, ResNet \cite{he2016deep} can be exploited. Compared with voxel-based models, these methods can improve the performance for different 3D tasks by taking multi-view of the interest object or scenes and then fusing or voting the outputs for final prediction. Compared with the above three different 3D data representations, view-based models can achieve near-optimal results, as shown in Table \ref{t2}. Su et al. \cite{su2018deeper} experimented that multiview methods have the optimal generalization ability even without using pre-trained models compared with point cloud and voxel data representation models. The advantages of view-based models compared with 3D models can be concluded as:
\begin{itemize}
    \item Efficiency. Compared with 3D data representations such as point clouds or voxel grids, the reduced one dimension information can greatly reduce the computation cost but with increased resolution \cite{su2015multi}. 
    \item Exploiting established 2D deep architectures and datasets. The well-developed 2D DL architectures can better exploit the local and global information from projected 2D view images \cite{you2018pvnet}. Besides, existing 2D image databases (such as ImageNet \cite{russakovsky2015imagenet}) can be used to train 2D DL architectures. 
\end{itemize}

\textbf{Multi-View CNN} (MVCNN) \cite{su2015multi} is the pioneer in exploiting 2D DL models to learn 3D representation. Multiple views of 3D objects are extracted without specific order using a view pooling layer. Two different CNNs models are proposed and tested in this paper. The first CNN model takes 12 views rendered from the object via placing 12 virtual cameras with equal distance around the objects as the input, while the second CNN model takes 80 views rendered in the same way as input. These views are first learned separately and then fused through max-pooling operation the extract the most representative feature among all views for the whole 3D shape. This network is effective and efficient compared with volumetric data representation. However, the max-pooling operation only considers the most important views and discards information from other views, which fails to preserve comprehensive visual information.

\textbf{MVCNN-MultiRes} was proposed by Qi et al  \cite{qi2016volumetric} to improve multi-view CNNs. Different from traditional view rendering methods, the 3D shape is projected to 2D via a convolution operation based on an anisotropic probing kernel applied to the 3D volume. Multi-orientation pooling is combined together to improve the 3D structure capturing capability. Then the MVCNN \cite{su2015multi} is applied to classify the 2D projects. Compared with MVCNN \cite{su2015multi}, multi-resolution 3D filtering is introduced to capture multi-scale information. Sphere rendering is performed at different volume resolutions to achieve view-invariant and improve the robust to potential noise and irregularities. This model achieves better results in 3D object classification task compared with MVCNN \cite{su2015multi}.

\textbf{3DMV} \cite{dai20183dmv} combines the geometry and imagery data as input to train a joint 3D deep architecture. Feature maps extracted from imagery data are first extracted and then mapped into the 3D feature extracted from the volumetric grid data derived from a differentiable back-projection layer. Because there exists redundant information among multiple views, a multiview pooling approach is applied to extract useful information from these views. This network achieved remarkable results in 3D objects classification. However, compared with models using one source of data such as LiDAR point or RGB images solely, the computation cost of this method is higher. 

\textbf{RotationNet} \cite{kanezaki2018rotationnet} is proposed following the assumption that when the object is observed by a viewer from a partial set of full multiview images, the observation direction should be recognized to correctly infer the object’s category. Thus, the multiview images of an object are input to the RotationNet, which outputs its pose and category. The most representative characteristic of RotationNet is that it treats viewpoints which are the observation of training images as latent variables. Then unsupervised learning of object poses is conducted based on an unaligned object dataset, which can eliminate the process of pose normalization to reduce noise and individual variations in shape. The whole network is constructed as a differentiable MLP network with softmax layers as the final layer. The outputs are the viewpoint category probabilities, which correspond to the predefined discrete viewpoints for each input image. These likelihoods are optimized by the selected object pose. 

However, there some limitation of 2D view-based models: 
\begin{itemize}
    \item The first is that the projection from 3D space to 2D views can lose some geometrically-related spatial information.
    \item  The second is the redundant information among multiple views. 
\end{itemize}

\subsection{3D Data Processing and Augmentation}
Due to the massive amount of data and the tedious labeling process, there exist limited reliable 3D datasets. To better exploit the architecture of deep networks and improve the model generalization ability, data augmentation is commonly conducted. Augmentation can be applied to both data space and feature space, while the most common augmentation is conducted in the first space. This type of augmentation can not only enrich the variations of data but also can generate new samples by conducting transformations to the existing 3D data. There are several types of transformations, such as translation, rotation, and scaling. Several requirements for data augmentation are summarised as: 
\begin{itemize}
    \item There must exist similar features between original augmented data, such as shape;
    \item There must exist different features between original and augmented data such as orientation.
\end{itemize}

Based on those existing methods, classical data augmentation for point clouds can be concluded as:
\begin{itemize}
    \item Mirror $x$ and $y$ axis with predefined probability \cite{roynard2018classification,zhi2018toward}
    \item Rotation around z-axis with certain times and angles\cite{roynard2018classification,zhi2017lightnet,zhi2018toward,zhou2018voxelnet}
    \item Random (uniform) height or position jittering in certain range \cite{dai2018scancomplete,zhi2018toward,maturana2015voxnet}
    \item Random scale with certain ratio \cite{roynard2018classification,zhou2018voxelnet}
    \item Random occlusions or randomly down-sampling points within predefined ratio \cite{roynard2018classification}
    \item Random artefacts or randomly down-sampling points within predefined ratio \cite{roynard2018classification}
    \item Randomly adding noise, following certain distribution, to the points' coordinates and local features  \cite{roynard2018classification,li2018so,luo2019learning}.
\end{itemize}

\section{Deep Learning in LiDAR Point Cloud for AVs}
The application of LiDAR point clouds for AVs can be concluded into three types: 3D point cloud segmentation, 3D object detection and localization, and 3D objects classification and recognition. Targets for these tasks vary, for example, scene segmentation focus on per-point label prediction, while detection and classification concentrate on integrated point set labeling. But they all need to exploit the input point feature representations before feature embedding and network construction.

We first make a survey of input point cloud feature representations applied in DL architectures for all these three tasks, such as local density and curvature. These features are representations of a specific 3D point or position in 3D space, which describe the geometrical structures and features based on the extracted information around the point. These features can be grouped into two types: one is derived directly from the sensors such as coordinate and intensity, we term them as direct point feature representations; the second is extracted from the information provided by each point’s neighbors, we term them as geo-local point feature representations.

\subsubsection{Direct input point feature representations}
The direct input point feature representations are mainly provided by laser scanners, which include the $x$, $y$, and $z$ coordinates, and other characteristics (e.g., intensity, angle, and number of returns). Two most frequently used features applied in DL are selected:
\begin{itemize}
    \item \textbf{XYZ coordinate}. The most direct point feature representation is the $XYZ$ coordinate provided by the sensors, which means the position of a point in the real world coordinate. 
    \item \textbf{Intensity}. The intensity represents the reflectance characteristics of the material surface, which is one common characteristic of laser scanners \cite{huang2017traffic}. Different objects have different reflectance, thus produce different densities in point clouds. For example, traffic signs have a higher intensity than vegetation. 
\end{itemize}

\subsubsection{ Geo-local point feature representations}
Local input point feature embeds the spatial relationship of points and their neighborhoods, which plays a significant role in point cloud segmentation \cite{qi2017pointnet++}, object detection \cite{qi2019votenet}, and classification \cite{wang2018dynamic}. Besides, the searched local region can be exploited by some operations such as CNNs \cite{lei2018spherical}. Two most representative and widely-used neighborhood searching methods are k-nearest neighbors (KNN) \cite{qi2017pointnet++,li2018so, engelmann2018know} and spherical neighborhood \cite{weinmann2015semantic}.

The geo-local feature representations are usually generated from the searched region using the above two neighborhood searching algorithms. They are composed of eigenvalues (e.g., ${\eta}_0$, ${\eta}_1$ and ${\eta}_2$ (${\eta}_0 > {\eta}_1 > {\eta}_2$)) or eigenvectors (e.g., $\overrightarrow{v_0}$, $\overrightarrow{v_1}$, and $\overrightarrow{v_2}$) by decomposing the covariance matrix defined in the searched region. We list five most commonly used 3D local feature descriptors applied in DL: 
\begin{itemize}
\item \textbf{Local density}. The local density is typically determined by the quantity of points in a selected area \cite{che2017fast}. Typically, the point density decreases when the distance of objects to the LiDAR sensor increases. In voxel-based models, the local density of points is related to the setting of voxel sizes \cite{vo2015octree}. 
\item \textbf{Local normal}. It infers the direction of the normal at a certain point on the surface. The equation about normal extraction can be found in \cite{zhang2018large}. In \cite{rusu2011point}, the eigenvector $\overrightarrow{v_2}$ of ${\eta}_2$ in $C_i$  is selected as the normal vector for each point. However, in \cite{qi2017pointnet}, the eigenvectors of  ${\eta}_0$, ${\eta}_1$ and ${\eta}_2$ are all chose as the normal vectors of point $p_i$. 
\item \textbf{Local curvature}. The local curvature is defined to be the rate at which the unit tangent vector changes direction. Similar to local normal calculation in \cite{zhang2018large}, the surface curvature change in \cite{rusu2011point} can be estimated from the eigenvalues derived from the Eigen decomposition: 
$curvature={\eta}_0/({\eta}_0+{\eta}_1+{\eta}_2)$
\item \textbf{Local linearity}. It is a local geometric characteristic for each point to indicate the linearity of its local geometry \cite{thomas2018semantic}: $linearity=\left(\eta_{1}-\eta_{2}\right) / \eta_{1}$. 
\item \textbf{Local planarity}. It describes the flatness of a given point neighbors. for example, group points have higher planarity compared with tree points \cite{thomas2018semantic}: $planarity = \left(\eta_{2}-\eta_{3}\right) / \eta_{1}$
\end{itemize}

\subsection{LiDAR point cloud semantic segmentation}
\begin{figure*}[!t]
\centering
\includegraphics[scale=0.45]{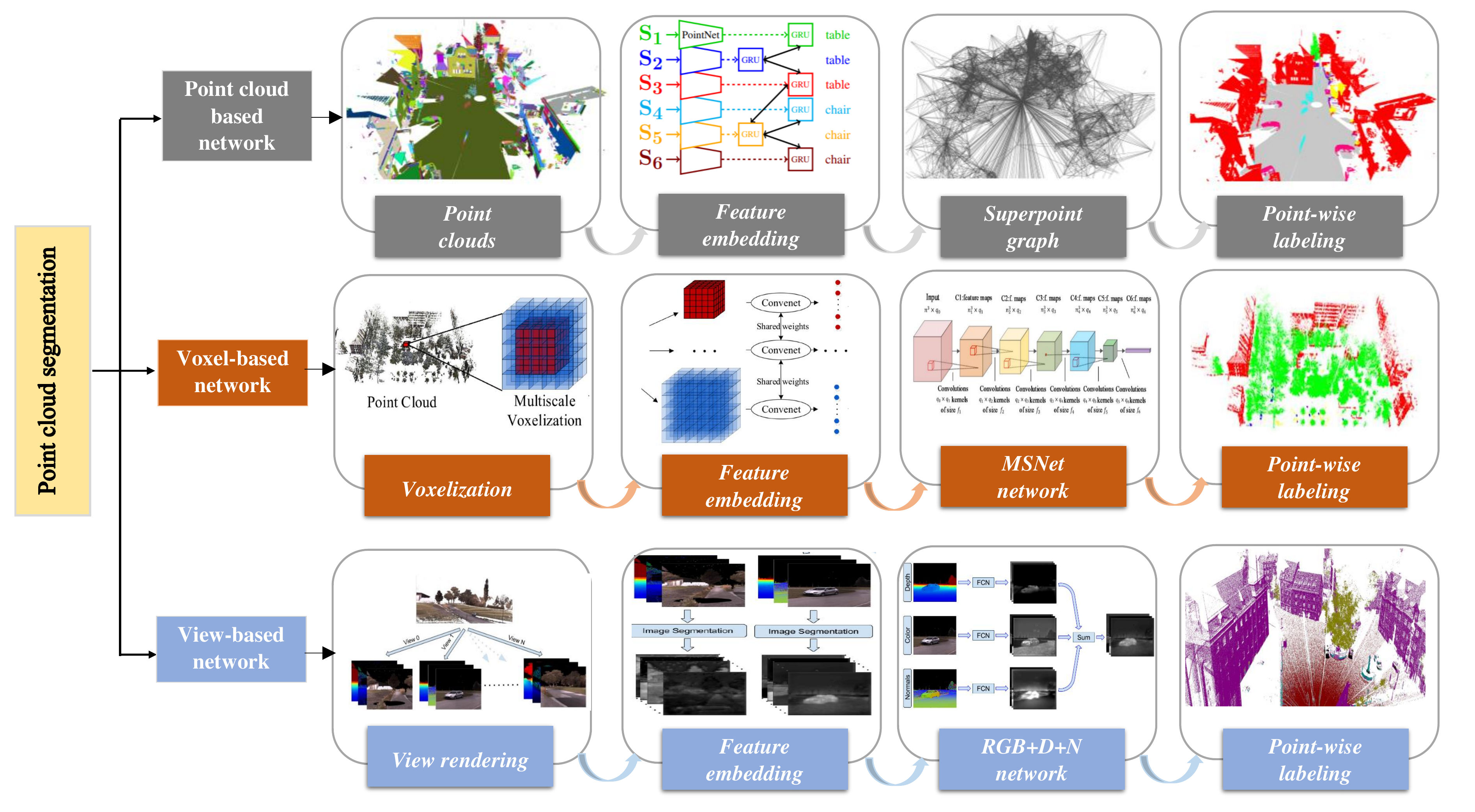}
\caption{DL architectures on LiDAR point cloud segmentation with three different data representations: point cloud based networks represented by SPG \cite{landrieu2018large}, voxel-based networks represented by MSNet \cite{wang2018msnet}, view-based networks represented by DeePr3SS \cite{lawin2017deep}}.
\label{fig_seg}
\end{figure*}

The goal of semantic segmentation is to label each point as belonging to a specific semantic class. For AVs segmentation tasks, these classes cloud be a street, buildings, cars, pedestrians, trees or traffic lights. When applying DL for point cloud segmentation, classification of small features is required \cite{treml2016speeding}. However, the LiDAR 3D point clouds are usually acquired in large scale, and they are irregularly shaped with changeable spatial contents. In the review of the recent five years papers related in this region, we group these papers into three schemes according to the types of data representation: point cloud based, voxel-based, and multi-view based models. There is limited research focusing on graph-based models, thus we combine the graph-based and point cloud based models together to illustrate their paradigms. Each type of model is represented by a compelling deep architecture as shown in Fig.\ref{fig_seg}. 

\subsubsection{Point cloud based networks}
For point cloud based networks, they are mainly composed of two parts: feature embedding and network construction. For the discriminate feature representing, both local and global features have demonstrated to be crucial for the success of CNNs \cite{qi2017pointnet++}. However, in order to apply conventional CNNs, the permutation and orientation problem for unordered and unoriented points requires a discriminative feature embedding network. Besides, lightweight, effective, and efficient deep network construction is another key module that affects the segmentation performance.

Local feature is commonly extracted from points neighborhoods \cite{thomas2018semantic}. The most frequently used local features are local normal and curvature \cite{qi2017pointnet, qi2017pointnet++}. To improve the receptive field, PointNet \cite{qi2017pointnet} has been proved to be a compelling architecture to extract semantic feature from unordered point sets. Thus, in \cite{qi2017pointnet++, rethage2018fully, landrieu2018large,wang2018sgpn}, a simplified PointNet is exploited to abstract local features from sampled point sets into high-level representations. Landrieu et al. \cite{landrieu2018large} proposed superpoint graph (SPG) to represent large 3D point clouds as a set of interconnected simple shapes coined superpoints, then PointNet is operated on these superpoints to embed features. 

To solve the permutation problem and extract local features, Huang et al. \cite{huang2018recurrent} proposed a novel slice pooling layer to extract the local context layer from the input point features and outputs an ordered sequence of aggregated features. To this end, the input points are first grouped into slices and then a global representation for each slice is generated via concatenating points features within the slice. The advantage of this slice pooling layer is the low computation cost compared with point-based local features. However, the slice size is sensitive to the density of data. In \cite{su2018splatnet}, bilateral Conv layers (BCL) are applied to perform convolutions on occupied parts of the lattice for hierarchical and spatially-aware feature learning. BCL first maps input points onto a sparse lattice and applies convolutional operations on the sparse lattice and then the filtered signal are interpolated smoothly to recover the original input points.

To reduce the computation cost, in \cite{rethage2018fully}, an encoding-decoding framework is adopted. Features extracted from the same scale of abstraction are combined and then upsampled by 3D deconvolutions to generate the desired output sampling density, which is finally interpolated by Latent nearest-neighbor interpolation to output per-point label. However, the down-sampling and up-sampling operations are hard to preserve the edge information, thus cannot extract the fine-grained features. In \cite{huang2018recurrent}, RNNs are applied to model dependencies of the ordered global representation derived from slice pooling. Similar to sequence data, each slice is viewed as one timestamp and the interaction information with other slices also follows the timestamps in RNN units. This operation enables the model to generate dependencies between slices. 

Although Zhang et al. \cite{zhang2018large} proposed the ReLu-NN to learn embedded point features, which is a four-layer MLP architecture. However, for objects without discriminative features, such as shrubs or trees, their local spatial relationship is not fully exploited. To better leverage the rich spatial information of objects, Wang et al. constructed a lightweight and effective deep neural network with spatial pooling (DNNSP) \cite{wang2018deep} to learn point features. They clustered the input data into groups and then applied distance minimum spanning tree-based pooling to extract the spatial information among the points in the clustered point sets. Finally, an MLP is used for classification with these features. In order to achieve multiple tasks, such as instance segmentation and object detection with simple architecture, Wang et al. \cite{wang2018sgpn} proposed a similarity group proposal network SGPN. Within the extracted local and global point features by PointNet, feature extraction network generates a matrix which is then diverged into three subsets that each pass through a single PointNet layer to obtain three similarity matrices. These three matrices are used to produce a similarity matrix, a confidence map and a semantic segmentation map. 

\subsubsection{Voxel-based networks}
In voxel-based networks, the point clouds are first voxelized into grids and then learn features from these grids. The deep network is finally constructed to map these features into segmentation masks.

Wang et al. \cite{wang2018msnet} conducted a multi-scale voxelization method to extract objects’ spatial information at different scales to form a comprehensive description. At each scale, a neighboring cubic with selected length is constructed for a given point \cite{huang2016point}. After that, the cube is divided into grid voxels with different size as a patch. The smaller the size is, the finer the scale. The point density and occupancy are selected to represent each voxel. The advantage of this kind voxelization is that it can accommodate objects with different sizes without losing their spatial space information. In \cite{tchapmi2017segcloud}, the class probabilities for each voxel are predicted using 3D-FCNN, which are then transferred back to the raw 3D points based on trilinear interpolation. In \cite{wang2018msnet}, after the multi-scale voxelization of point clouds,  features at different scales and spatial resolutions are learned by a set of CNNs with shared weights which are finally fused together for final prediction.

In voxel-based point cloud segmentation task, there are two ways to label each point: (1) Using the voxel label derived from the argmax of the predicted probabilities; (2) Further globally optimizing the class label of the point cloud based on spatial consistency. The first method is simple, but the result is provided at the voxel level and inevitably influenced by noise. The second one is more accurate but complex with additional computation. Because the inherent invariance of CNN networks to spatial transformations affects the segmentation accuracy \cite{garcia2017review}. In order to extract the fine-grained details for volumetric data representations, the Conditional Random Field (CRF) \cite{lafferty2001conditional, tchapmi2017segcloud, wang2018msnet} is commonly adopted in a post-processing stage. The CRFs have the advantage in combining low-level information such as the interactions between points to output multi-class inference for multi-class per-point labeling tasks, which compensates the fine local details that CNNs fail to capture.

\subsubsection{Multiview-based networks}

As for multi-view based models, view rendering and deep architecture construction are two key modules for segmentation task. The first one is used to generate structural and well-organized 2D grids that can exploit existing CNN-based deep architectures. The second one is proposed to construct the most suitable and generative models for different data. 

In order to extract local and global features simultaneously, some hand-designed feature descriptors are employed for representative information extraction. In \cite{zhang2018large,wang2018deep}, the spin image descriptor is employed to represent point-based local features, which contains the global description of objects from partial views and clutters of local shape description. In \cite{lawin2017deep}, point splatting was applied to generate view images by projecting the points with a spread function into the image plane. The point is first projected into image coordinates of a virtual camera. For each projected point, its corresponding depth value and feature vectors such as normal are stored. 

Once the points are projected into multi-view 2D images, some discriminative 2D deep networks can be exploited, such as VGG16 \cite{simonyan2014very}, AlexNet \cite{krizhevsky2012imagenet}, GoogLeNet \cite{szegedy2015going}, and ResNet \cite{he2016deep}. In \cite{garcia2017review}, these deep networks have been detailed analyzed in 2D semantic segmentation. Among these methods, VGG16 \cite{simonyan2014very}, composed of 16 layers, is the most frequently used. Its main advantage is the use of stacked Conv layers with small receptive fields, which produces a lightweight network with limited parameters and increasing nonlinearity \cite{garcia2017review,zhang2018fusion,lawin2017deep}. 

\begin{figure*}[!t]
\centering
\includegraphics[scale=0.45]{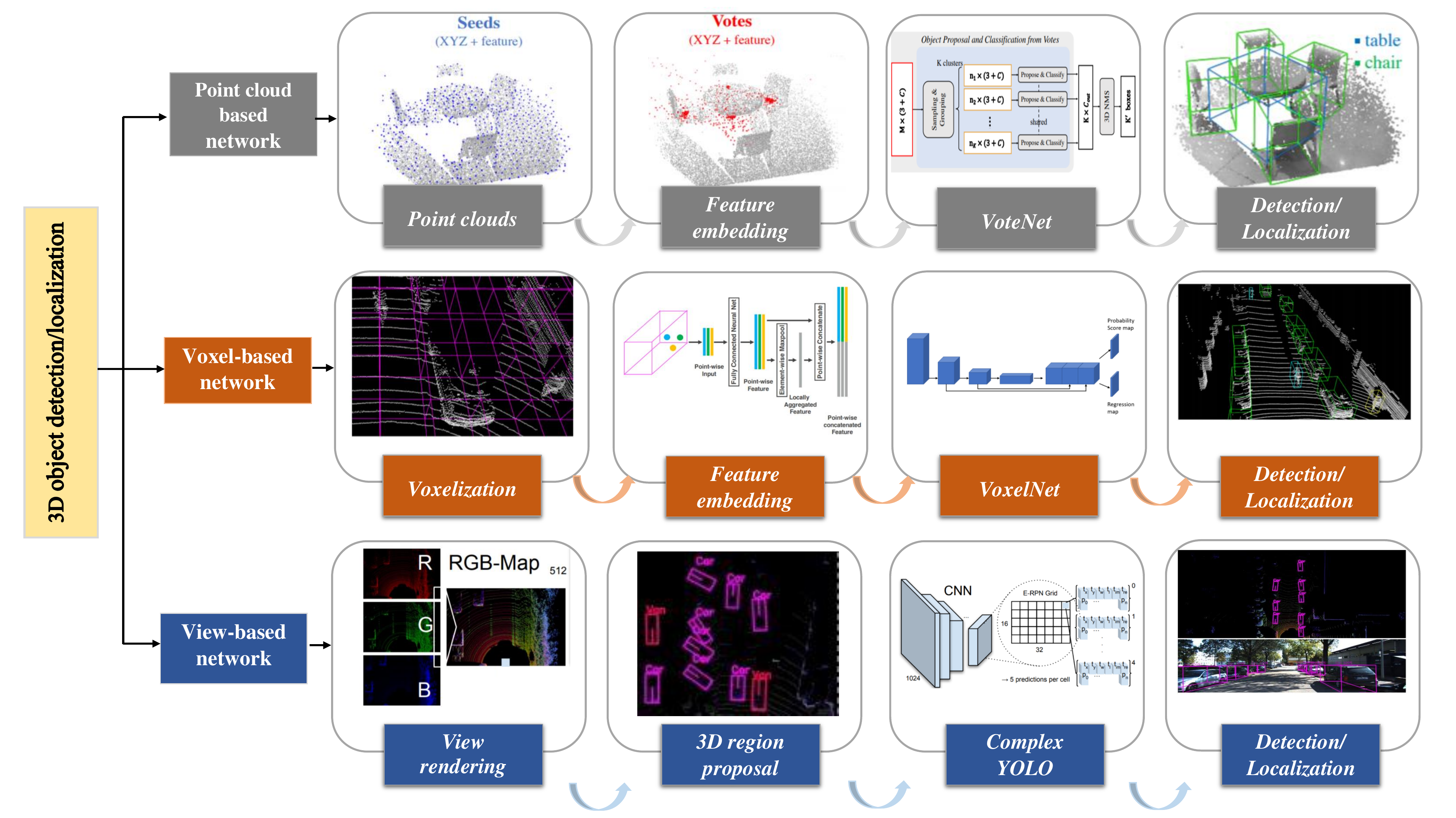}
\caption{DL architectures on 3D object detection/localization with three different data representations: point cloud based networks represented by VoteNet \cite{qi2019votenet}, voxel-based networks represented by VoxelNet \cite{zhou2018voxelnet}, view-based networks represented by ComplexYOLO \cite{simony2018complex}.}
\label{fig_detect}
\end{figure*}

\subsubsection{Evaluation on Point cloud segmentation}
Due to the high volume of point clouds, which pose a great challenge for computation capability. We choose the models tested on Reduced-8 Semantic3D dataset to compare their performance, as shown in Table \ref{t_seg}. Reduced-8 shares the same training
data as semantic-8 but only use a small part of test data, which can also suit the high computation cost algorithm for competing. The metrics used to compare these models are $IoU_{i}$, $\overline{IoU}$, and $\mathrm{OA}$. The computation efficiency for these algorithms are not reported and compared due to the difference between computation capacity, selected training dataset, model architecture.

\begin{table*}[]
\centering
\caption{Segmentation results on Semantic3D reduced-8 dataset}
\begin{tabular}{|l|l|l|l|l|l|l|l|l|l|l|l|l|c|}
\hline
\textbf{Method} & \textbf{Input} & \textbf{Backbone} & \multicolumn{8}{c|}{\textbf{IoU}} & \multicolumn{1}{c|}{\textbf{mIoU}} &\begin{tabular}[c]{@{}c@{}} \textbf{OA}\\ (\%)\end{tabular}&\textbf{Highlights}  \\ \cline{4-11}
 &  &  & \tiny{IoU1} & \tiny{IoU2} & \tiny{IoU3} & \tiny{IoU4} & \tiny{IoU5} & \tiny{IoU6} & \tiny{IoU7} & \tiny{IoU8} & \multicolumn{1}{c|}{} &  &  \\ \hline
\begin{tabular}[c]{@{}l@{}}SPGraph\\ \cite{landrieu2018large} \end{tabular} & \begin{tabular}[c]{@{}l@{}}Graph \\ \& \\ point \\ cloud\end{tabular} & \begin{tabular}[c]{@{}l@{}}PointNet\\ \cite{qi2017pointnet} \end{tabular} & \tiny{0.974} & \tiny{0.926} & \tiny{0.879} & \tiny{0.44} & \tiny{0.932} & \tiny{0.31} & \tiny{0.635} & \tiny{0.762} & 0.732 & 94.0 & \begin{tabular}[c]{@{}l@{}}3D point clouds are represented \\as a set of interconnected \\superpoints; Solve the big data \\ challenge and the unevenly \\distributed point density problem.\end{tabular} \\ \hline
\begin{tabular}[c]{@{}l@{}}MSDeepVoxNet\\ \cite{roynard2018classification}\end{tabular}& voxels & \begin{tabular}[c]{@{}l@{}}VGG16\\ \cite{simonyan2014very} \end{tabular} & \tiny{0.83} & \tiny{0.672} & \tiny{0.838} & \tiny{0.367} & \tiny{0.924} & \tiny{0.313} & \tiny{0.500} & \tiny{0.782} & 0.653 & 88.4 & \begin{tabular}[c]{@{}l@{}} Extract multi-scale local features \\in a multi-scale neighborhood.\end{tabular} \\ \hline
\begin{tabular}[c]{@{}l@{}}RF\_MSSF\\ \cite{thomas2018semantic}\end{tabular} & \begin{tabular}[c]{@{}l@{}}point  \\ cloud\end{tabular} & \begin{tabular}[c]{@{}l@{}}Random\\ forest \cite{liaw2002classification} \end{tabular} & \tiny{0.876} & \tiny{0.803} & \tiny{0.818} & \tiny{0.364} & \tiny{0.922} & \tiny{0.241} & \tiny{0.426} & \tiny{0.566} & 0.627 & 90.3 & \begin{tabular}[c]{@{}l@{}}Define multi-scale neighborhoods \\in point clouds to extract features.\end{tabular} \\ \hline
\begin{tabular}[c]{@{}l@{}}SEGCloud\\ \cite{tchapmi2017segcloud}\end{tabular}  & voxels & FCNN & \tiny{0.839} & \tiny{0.66} & \tiny{0.86} & \tiny{0.405} & \tiny{0.911} & \tiny{0.309} & \tiny{0.275} & \tiny{0.643} & 0.613 & 88.1 & \begin{tabular}[c]{@{}l@{}}Refine the labels generated at the \\voxel level for each point using\\ Trilinear Interpolation; a FC CRF \\is connected with the FCNN to \\improve the segmentation result.\end{tabular} \\ \hline
\begin{tabular}[c]{@{}l@{}}SnapNet\\ \cite{boulch2017unstructured}\end{tabular} & images & CNN & \tiny{0.82} & \tiny{0.773} & \tiny{0.797} & \tiny{0.229} & \tiny{0.911} & \tiny{0.184} & \tiny{0.373} & \tiny{0.644} & 0.591 & 88.6 & \begin{tabular}[c]{@{}l@{}}Generate RGB and depth images \\from point cloud; Use CNN to\\ conduct a pixel-wise labeling of\\ each pair of 2D snapshot; \\Back-projection of the label \\predictions in the 3D space.\end{tabular} \\ \hline
\begin{tabular}[c]{@{}l@{}}DeePr3SS\\ \cite{lawin2017deep}\end{tabular} & images & \begin{tabular}[c]{@{}l@{}}VGG16\\ \cite{simonyan2014very} \end{tabular}  & \tiny{0.856} & \tiny{0.832} & \tiny{0.742} & \tiny{0.324} & \tiny{0.897} & \tiny{0.185} & \tiny{0.251} & \tiny{0.592} & 0.585 & 88.9 & \begin{tabular}[c]{@{}l@{}}Generate view images from point \\clouds with a spread function \\into the image plane. Depth value \\and feature vectors of each \\projected point are stored and \\input to  VGG16 for segmentation\end{tabular} \\ \hline
\end{tabular}
\label{t_seg}
\end{table*}

\subsection{3D objects detection (localization)}
The detection(\& localization) of 3D objects in LiDAR point clouds can be summarised as bounding box prediction and objectness prediction \cite{li20173d}. In this paper, we mainly survey the LiDAR-only paradigm, which takes advantage from accurate geo-referenced information. Overall, there are two ways for data representation in this paradigm: one detects and locates 3D objects directly from point clouds \cite{zhang2018deep}; another first converts 3D points into regular grids, such as voxel grids or bird’s eye view images as well as front views, and then utilizes architectures in 2D detectors to extract object from images, the 2D detection results are finally back-projected into 3D space for final 3D object location estimation \cite{pang20163d}. Fig.\ref{fig_detect} shows the representative network frameworks of the above-listed data representations.

\subsubsection{3D objects detection (localization) from point clouds}
The challenges for 3D object detection from sparse and large-scale point clouds are concluded as:
\begin{itemize}
    \item The detected objects only occupy a very limited amount of the whole input data.
    \item The 3D object centroid can be far from any surface point thus hard to regress accurately in one step \cite{qi2019votenet}.
    \item The missing of 3D object center points. As LiDAR sensors only capture surfaces of objects, 3D object centers are likely to be in empty space, far away from any point.
\end{itemize}
Thus, a common procedure of 3D object detection and localization from large-scale point clouds is composed of the following processes: firstly, the whole scene is roughly segmented, and then the coarse location of interest object is approximately proposed; secondly, the feature for each proposed region is extracted; finally, the localization and object class is predicted through a Bounding-Box Prediction Network \cite{zhang2018deep,yang2018ipod}.

In \cite{yang2018ipod}, the PointNet++ \cite{qi2017pointnet++} is applied to generate per-point feature within the whole input point clouds. Different from \cite{zhang2018deep}, each point is viewed as an effective proposal, which preserves the localization information. Then the localization and detection prediction is conducted based on the extracted point-based proposal features as well as local neighbor context information captured by increasing receptive field and input point features. This network preserves more accurate localization information but has higher computation cost for operating directly on point sets. 

In \cite{zhang2018deep}, 3D CNN with three Conv layers and multiple FC layers is applied to learn the discriminate and robust features of objects. Then an intelligent eye window (EW) algorithm is applied to the scene. The label of point belong to the EW is predicted using the pre-trained 3D CNN. The evaluation result is then input to the deep Q-network (DQN) to adjust the size and position of EW. Then the new EW is evaluated by 3D CNN and DQN until the EW only contains one object. Different from the traditional bounding box of the region of interest (RoI), the EW can reshape its size and change the window center automatically, which is suitable for objects with different scales. Once the position of the object is located, the object in the input window is predicted with learned features. In \cite{zhang2018deep}, the object features are extracted based on 3D CNN models and then fed into the residual RNN \cite{zoph2018learning} for category labeling.

Qi et al. \cite{qi2019votenet} proposed VoteNet a 3D object detection deep network based on Hough voting. The raw point clouds are input to PointNet++ \cite{qi2017pointnet++} to learn point features. Based on these features, a group of seed points is sampled and generate votes from their neighbor features. These seeds are then gathered to cluster the object centers and generate bounding box proposals for a final decision. Compared with the above two architectures, VoteNet is robust to sparse and large-scale point clouds. Besides, it can localize the object center with high accuracy.

\subsubsection{3D objects detection (localization) from regular voxel grid} To better exploit CNNs, some approaches voxelize the 3D space into a voxel grid, which is represented by a scalar value such as occupancy or vector data extracted from voxels \cite{yang2018pixor}. In \cite{wang2015voting,engelcke2017vote3deep}, the 3D space is first discretized into grids with a fixed size and then converted each occupied cell into a fixed-dimensional feature vector. Non-occupied cells without any points are represented with zero feature vectors. A binary occupancy and the mean and variance of the reflectance, as well as three shape factors are used to describe the feature vector. For simplicity, in \cite{li20173d}, the voxelized grids are represented by length, width, height, and channels 4D array, and the binary value of one channel is used to represent the observation status of points in corresponding grids. Zhou et al. \cite{zhou2018voxelnet} voxelized the 3D point clouds along $XYZ$ coordinates with predefined distance and grouped points in each grid. Then a voxel feature encoding (VFE) layer is proposed to achieve inter-point interaction within a voxel, by combining per-point features and local neighbor features. The combination of multi-scale VFE layers enables this architecture to learn discriminative features from local shape information.

The voting scheme is adopted in \cite{wang2015voting,engelcke2017vote3deep} to perform a sparse convolution on the voxelized grids. These grids, weighted by the convolution kernels as well as their surrounding cells in the receptive field, accumulate the votes from their neighbors by flipping the CNN kernel along each dimension and finally outputs the voting scores for potential interest objects. Based on that voting scheme, Engelcke et al. \cite{engelcke2017vote3deep} then used a ReLU non-linearity to produce a novel sparse 3D representation of these grids. This process is iterated and stacked in conventional CNN operations and finally output the predicting scores for each proposal. However, the voting scheme has high computation during voting. Thus, modified region proposal networks (RPN) is employed by \cite{zhou2018voxelnet} in object detection to reduce computation. This RPN is composed of three blocks of Conv layers, which are used to downsample, filter features and upsample the input feature map and produce a probability score map, and a regression map for object detection and localization.

\subsubsection{3D objects detection (localization) from 2D views}
Some approaches also project LiDAR point clouds into 2D views. Such approaches are mainly composed of those two steps: first is the projection of 3D points; second is the object detection from projected images. There are several types of view generation methods to project 3D points into 2D images: BEV images  \cite{beltran2018birdnet,chen2017multi,ku2018joint,simony2018complex}, front view images \cite{chen2017multi}, spherical projections \cite{pang20163d}, and cylindral projection \cite{li2016vehicle}.

Different from \cite{pang20163d}, in \cite{beltran2018birdnet,chen2017multi,ku2018joint,simony2018complex}, the point cloud data is split into grids with fixed size and then converted to a bird’s eye view (BEV) image with corresponding three channels which encodes height, intensity, and density information. Considering the efficiency and performance, only the maximum height, the maximum intensity, and the normalized density among the grids are converted to a single birds-eye-view RGB-map \cite{simony2018complex}. In \cite{yu2017vehicle}, only the maximum, median, and minimum height values are selected to represent the channels of the BEV image to exploit conventional 2D RGB deep models without modification. Dewan et al. \cite{dewan2017deep} selected the range, intensity, and height values to represent three channels. In \cite{yang2018pixor}, the feature representation for each BEV pixel is composed of occupancy and reflectance value.

However, due to the sparsity of point clouds, the projection of point clouds to the 2D image plane produces a sparse 2D point map. Thus, Chen et al. \cite{chen2017multi} added front view representation to compensate for the missing information in BEV images. The point clouds are projected to a cylinder plane to produce dense front view images. In order to keep the 3D spatial information during projection, points are projected at multiview angles which are evenly selected on a sphere \cite{pang20163d}. Pang et al. first discretized 3D points into cells with a fixed size. Then the scene is sampled to generate multiview images to construct positive and negative training samples. The benefits of this kind of dataset generation are that the spatial relationship and feature of the scene can be better exploited. However, this model is not robust to a new scene and cannot learn new features from a constructed dataset. 
	
As for 2D object detectors, there exist enormous compelling deep models such as VGG-16 \cite{simonyan2014very}, Faster R-CNN \cite{ren2015faster}. In \cite{liu2018deep}, a comprehensive survey of 2D detectors in object detection is concluded. 

\subsubsection{Evaluation on 3D objects localization and detection}
In order to compare 3D objects localization and detection deep models, KITTI bird’s eye view benchmark and KITTI 3D object detection benchmark \cite{geiger2012we} are selected. As reported in \cite{geiger2012we}, all non- and weakly-occluded $(<20 \%)$ objects which are neither truncated nor smaller than 40 px in height are evaluated. Truncated or occluded objects are not counted as false positives. Only a bounding box overlap of at least $50\%$ results for pedestrian and cyclist, and $70\%$ results for the car are considered for detection, localization, and orientation estimation measurements. Besides, this benchmark classified the difficulties of tasks into three types: easy, moderate, and hard. 

Both the accuracy and execution time are compared to evaluate these algorithms because detection and localization in real-time are crucial for AVs \cite{zeng2018rt3d}. For the localization task, the KITTI bird’s eye view benchmark is chosen as the evaluation benchmark, and the comparison results are shown in Table \ref{t_detection}. The 3D detection is evaluated on the KITTI 3D object detection benchmark. Table \ref{t_detection} shows the runtime and the average precision ($AP_{3D}$) on the validation set. For each bounding box overlap, only 3D IoU exceeds 0.25 is considered as a valid localization/detection box \cite{zeng2018rt3d}.

\begin{table*}[]
\caption{3D car localization performance on KITTI bird’s eye view benchmark: average precision ($AP_{loc}$ $[\%]$)}
\centering
\resizebox{\textwidth}{45mm}{
\begin{tabular}{|l|l|l|l|l|l|l|l|l|l|l|l|l|l|l|l|l|l|l|c|}\hline
\textbf{Method}& \textbf{Input} &\begin{tabular}[c]{@{}c@{}} \textbf{Times} \\  (s) \end{tabular} &\begin{tabular}[c]{@{}l@{}} \textbf{GPUs}\end{tabular} & \multicolumn{15}{c|}{\textbf{Evaluation on AP }(\%)} &{\textbf{Highlights}} \\ \cline{5-19}
 &  & \multicolumn{1}{c|}{} &  & \multicolumn{9}{c|}{\textbf{object detection}} & \multicolumn{6}{c|}{\textbf{object localization}} & \multicolumn{1}{c|}{} \\ \cline{5-19}
 &  & \multicolumn{1}{c|}{} &  & \multicolumn{3}{c|}{0.25} & \multicolumn{3}{c|}{0.5} & \multicolumn{3}{c|}{0.7} & \multicolumn{3}{c|}{0.5} & \multicolumn{3}{c|}{0.7} & \multicolumn{1}{c|}{} \\ \cline{5-19}
 &  & \multicolumn{1}{c|}{} &  & \tiny{E }& \tiny{M }& \tiny{H} & \tiny{E} & \tiny{M} & \tiny{H} & \tiny{E} & \tiny{M} & \tiny{H} & \tiny{E} & \tiny{M} & \tiny{H} & \tiny{E} & \tiny{M} & \tiny{H} & \multicolumn{1}{c|}{} \\ \hline
\begin{tabular}[c]{@{}l@{}}VeloFCN\\ \cite{li20173d}\end{tabular} & voxels & 1 & \multicolumn{1}{c|}{N/A}& \tiny{89.0} & \tiny{81.1} & \tiny{75.9} & \tiny{67.9} & \tiny{57.6} & \tiny{52.6} & \tiny{15.2} & \tiny{13.7} & \tiny{16.0} & \tiny{79.7} & \tiny{63.8} & \tiny{62.8} & \tiny{40.1} & \tiny{32.1} & \tiny{30.5} & \begin{tabular}[c]{@{}l@{}}The voxelized grids are represented \\ by length, width, height, and channels; \\ The binary value of one channel is \\used to represent the observation\\ status of points in corresponding grids.\end{tabular} \\ \hline
\begin{tabular}[c]{@{}l@{}}DOBEM\\ \cite{yu2017vehicle}\end{tabular} & images & 0.6 & \multicolumn{1}{c|}{\begin{tabular}[c]{@{}c@{}}Titan\\  X\end{tabular}} & \tiny{N/A} & \tiny{N/A} & \tiny{N/A} & \tiny{N/A} & \tiny{N/A} & \tiny{N/A} & \tiny{N/A} & \tiny{N/A} & \tiny{N/A} & \tiny{79.3} & \tiny{80.2} & \tiny{80.1} & \tiny{54.9} & \tiny{60.1} & \tiny{60.9} & \begin{tabular}[c]{@{}l@{}}The maximum, median and minimum \\ height values are selected to represent \\ the BEV image channels to \\ exploit conventional 2D RGB deep \\ models without modification.\end{tabular} \\ \hline
\begin{tabular}[c]{@{}l@{}}MV3D\\ \cite{chen2017multi}\end{tabular} & images & 0.36 & \multicolumn{1}{c|}{\begin{tabular}[c]{@{}c@{}}Titan\\  X\end{tabular}}& \tiny{96.5} & \tiny{89.6} & \tiny{88.9} & \tiny{96.0} & \tiny{89.1} & \tiny{88.4} & \tiny{71.3} & \tiny{62.7} & \tiny{56.6} & \tiny{96.3} & \tiny{89.4} & \tiny{88.7} & \tiny{86.6} & \tiny{78.1} & \tiny{76.7} & \begin{tabular}[c]{@{}l@{}}The point clouds are projected to a\\ cylinder plane to produce dense\\  front view images\end{tabular} \\ \hline
\begin{tabular}[c]{@{}l@{}}VoxelNet\\ \cite{zhou2018voxelnet}\end{tabular} & voxels & 0.23 & \multicolumn{1}{c|}{\begin{tabular}[c]{@{}c@{}}Titan\\  X\end{tabular}} & \tiny{N/A} & \tiny{N/A} & \tiny{N/A} & \tiny{N/A} & \tiny{N/A} & \tiny{N/A} & \tiny{82.0} & \tiny{65.5} & \tiny{62.9} & \tiny{N/A} & \tiny{N/A} & \tiny{N/A} & \tiny{89.6} & \tiny{84.8} & \tiny{78.6} & \begin{tabular}[c]{@{}l@{}}Voxelize the 3D point clouds along \\ XYZ coordinates with predefined\\  distance and group points in each grid; \\ Then a voxel feature encoding \\ layer is proposed to achieve \\ inter-point interaction within a voxel, \\ by combining per-point features and \\ local neighbor features.\end{tabular} \\ \hline
\begin{tabular}[c]{@{}l@{}}RT3D\\ \cite{zeng2018rt3d}\end{tabular} & \begin{tabular}[c]{@{}l@{}}point\\  cloud\end{tabular} & 0.09 & \multicolumn{1}{c|}{\begin{tabular}[c]{@{}c@{}}Titan\\  X\end{tabular}} & \tiny{89.5} & \tiny{81.0} & \tiny{81.2} & \tiny{89.0} & \tiny{80.6} & \tiny{80.9} & \tiny{72.9} & \tiny{61.6} & \tiny{64.4} & \tiny{89.4} & \tiny{80.9} & \tiny{81.2} & \tiny{88.3} & \tiny{79.9} & \tiny{80.4} & \begin{tabular}[c]{@{}l@{}}Propose a pre-RoI-pooling convolution\\  technique that moves a majority of the \\ convolution operations to the RoI pooling.\end{tabular} \\ \hline
\end{tabular}}
\label{t_detection}
\end{table*}

\subsection{3D object classification}
Semantic object classification/recognition is crucial for safe and reliable driving of AVs in unstructured and uncontrolled real-world environments \cite{maturana2015voxnet}. Existing 3D object detection are mainly focus on CAD data (e.g., ModelNet40 \cite{wu20153d}) or RGBD data (e.g., NYUv2 \cite{silberman2012indoor}). However, these data have uniform point distribution, complete shapes, limited noise, occlusion and background clutter, which poses limit challenges for 3D classification compared with LiDAR point clouds \cite{qi2017pointnet, qi2017pointnet++, xu2018spidercnn}. Those compelling deep architectures applied on CAD data have been analyzed in the form of four types of data representations in section III. In this part, we mainly focus on the LiDAR data based deep models for the classification task.

\subsubsection{Volumetric architectures} The voxelization of point clouds depends on the data spatial resolution, orientation, and the origin \cite{maturana2015voxnet}. This operation which can provide enough recognizable information but not increase the computation cost is crucial for DL models. Thus, for LiDAR data, a voxel with spatial resolution such as $(0.1 m)^3$ is adopted in \cite{maturana2015voxnet} to voxelize the input points. Then for each voxel, binary occupancy grid, density grid, hit grid are calculated to estimate its occupancy. The input layer, Conv layer, pooling layer, and FC layer are combined to construct the CNNs. Such architecture can exploit the spatial structure among data and extract global feature via pooling. However, the FC layer produces high computation cost and lose the spatial information between voxels. In \cite{sedaghat2016orientation}, based on VoxNet \cite{maturana2015voxnet}, it takes a 3D voxel grid as input and contains two Conv layers with 3D filters followed by two FC layers. Different from other category-level classification tasks, they treated this task as a multi-task problem, where the orientation estimation and class label prediction are processed parallel.

For simplicity and efficiency, Zhi et al. \cite{zhi2018toward,ma2018binary} adopted the binary grid of \cite{maturana2015voxnet} to reduce the computation cost. However, they only consider the voxels inside the surface, ignoring the difference between unknown and free space. Normal vectors, which contain geo-local position and orientation information, have been demonstrated stronger than binary grid in \cite{wang2019normalnet} Similar to \cite{sedaghat2016orientation}, the classification is treated as two tasks: voxel object class label predicting and its orientation prediction. To extract local and global features, there are two sub-tasks in the first task: the first sub-task is to predict the object label referencing the whole input shape while the second one predicts the object label with part of the shape. The orientation prediction is proposed to exploit the orientation augmentation scheme. The whole network is composed of three 3D Conv layers and two 3D max-pooling layers, which is lightweight and demonstrated robust to occlusion and clutter.
	
\subsubsection{Multi-view architectures} The merit of view-based methods is their ability to exploit both local and global spatial relationships among points. Luo et al. \cite{luo2019learning} designed the three feature descriptors to extract local and global features from point clouds: the first one captures the horizontal geometric structure, the second one extracts vertical information, the last one provides complete spatial information. To better leverage the multi-view data representations, You et al. \cite{you2018pvnet} integrated the merits of point cloud and multi-view data and achieved better results than MVCNN \cite{su2015multi} in 3D classification. Besides, the high-level features extracted from view representations based on MVCNN \cite{su2015multi} are embedded with an attention fusion scheme to compensate the local features extracted from point cloud data representations. Such attention-aware features are proved efficient in representing discriminative information of 3D data.

However, for different objects, the view generation process varies. Because the special attributes of objects can contribute to computation saving and accuracy improving. For example, in road marking extraction tasks, the elevation derived mainly from $Z$ coordinate contributes little to the algorithm. But the road surface is actually a 2D structure. As a result, Wen et al. \cite{wen2019deep} directly projected 3D point clouds onto a horizontal plane and girded as a 2D image. Luo et al. \cite{luo2019learning} input the acquired three-view descriptors separately to capture low-level features to JointNet. Then this network learns high-level features by a convolutional operation based on the input features, and finally fuses the prediction scores. The whole framework is composed of five Conv layers, a spatial pyramid pooling (SPP) layer \cite{he2015spatial} and two FC layers and a reshape layer. The output results are fused through Conv layers and multi-view pooling layers. The well-designed view descriptors help the network achieve compelling results in object classification tasks. 

Another representative architecture in 2D deep models is the encoder-decoder architecture. Due to the down-sampling and up-sampling can help to compress the information among pixels to extract the most representative features. In \cite{wen2019deep}, Wen et al. proposed a modified U-net model to classify road markings. The point clouds data are first mapped into the intensity images. Then a hierarchical U-net module is applied to classify road markings by multi-scale clustering via CNNs. Due to such down-sampling and up-sampling is hard to preserve the fine-grained patterns, a GAN network is adopted to reshape small-size road markings, broken lane lines and missing marking considering the expert context knowledge. This architecture exploits the efficiency of U-net and completeness of GAN to classify the road markings with high efficiency and accuracy.

\subsubsection{Evaluation on 3D objects classification}
There is limited published LiDAR point cloud benchmark specific for 3D objects classification task. Thus, the Sydney Urban Objects dataset is selected due to the performance of several state-of-the-art methods are available. The $F_1$ score is used to evaluate these published algorithms \cite{luo2019learning}, as shown in Table \ref{t_classification}.

\begin{table}[]
\caption{3D classification performance on the Sydney Urban Objects dataset \cite{luo2019learning}}
\centering
\begin{tabular}{|l|c|c|l|}
\hline
\multicolumn{1}{|c|}{\textbf{Method}} & \textbf{Input} & \begin{tabular}[c]{@{}c@{}} \textbf{$F_1$ score}\\ {(\%)}\end{tabular} & \multicolumn{1}{c|}{\textbf{Highlights}} \\ \hline
\begin{tabular}[c]{@{}l@{}}VoxNet\\ \cite{maturana2015voxnet} \end{tabular} & voxels & 72.0 & \begin{tabular}[c]{@{}l@{}}The input points are voxelized \\with spatial resolution; Binary \\occupancy grid, density grid, \\hit grid are calculated to\\ estimate each voxel occupancy.\end{tabular} \\ \hline
\begin{tabular}[c]{@{}l@{}}BV-CNNs\\ \cite{ma2018binary}\end{tabular} & voxels & 75.5 & \begin{tabular}[c]{@{}l@{}}Transform the inputs and weights \\in FC layers to binary values,\\which can potentially accelerate \\the networks by bit-wise\end{tabular} \\ \hline
\begin{tabular}[c]{@{}l@{}}ORION\\ \cite{sedaghat2016orientation} \end{tabular} & voxels & 77.8 & \begin{tabular}[c]{@{}l@{}}Category level classification \\task is treated as a multitask \\problem, where the orientation \\ estimation and class label \\prediction are processed \\parallel.\end{tabular} \\ \hline
\begin{tabular}[c]{@{}l@{}}JointNet\\ \cite{luo2019learning} \end{tabular} & images & 74.9 & \begin{tabular}[c]{@{}l@{}}Three feature descriptors are\\ proposed to  extract local and \\global features from point\\ clouds: the horizontal geometric \\structure, vertical information, \\complete spatial information.\end{tabular} \\ \hline
\end{tabular}
\label{t_classification}
\end{table}

\section{Research Challenges and Opportunities}
DL architectures developed in recent five years using LiDAR point clouds have made significant success in the field of autonomous driving detailing for 3D segmentation, detection, and classification tasks. However, there still exists a huge gap between cutting-edge results and human-level performance. Although there is much work to be done, we mainly summarize the remaining challenges specific for data, deep architectures, and tasks as follows:

\subsubsection{\textbf{Multi-source Data Fusion}}
To compensate the absence of 2D semantic, textual and incomplete information in 3D points, imagery, LiDAR point clouds, and radar data can be fused to provide accurate, geo-referenced, and information-rich cues for AVs' navigation and decision making \cite{liang2018deep}. Besides, there also exists a fusion between data acquired by low-end LiDAR (e.g., Velodyne HDL-16E) and high-end LiDAR (e.g., Velodyne HDL-64E) sensors. However, there exist several challenges in fusing these data: The first is the sparsity of point clouds causes the inconsistent and missing data when fusing multi-source data. The second is that the existing data fusion scheme using DL knowledge is processed in a separate line, which is not an end-to-end scheme.  \cite{qi2018frustum,yang2018ipod,Xu_2018_PointFusion}.

\subsubsection{\textbf{Robust Data Representation}} 
The unstructured and unordered data format \cite{qi2017pointnet, qi2017pointnet++} poses a great challenge for robust 3D DL applications. Although there are several effective data representations such as voxels \cite{maturana2015voxnet}, point clouds \cite{qi2017pointnet,qi2017pointnet++}, graphs \cite{wang2018dynamic,xu2018spidercnn}, 2D views \cite{kanezaki2018rotationnet}, or novel 3D data representations \cite{He2019GeoNet,Mescheder2018Occupancy,Le2018PointGrid}, there has not yet agreed on a robust and memory-efficient 3D data representation. For example, although voxels solve the ordering problem, the computation cost increases cubically with the increment of voxel resolution \cite{wu20153d, maturana2015voxnet}. As for point clouds and graphs, the permutation invariance and the computation capability limit the processable quantity of points, which inevitably constrains the performance of the deep models \cite{qi2017pointnet,wang2018dynamic}. 

\subsubsection{\textbf{Effective and More Efficient Deep Frameworks}} 
Due to the limitation of memory and computation facilities of the platform embedded in AVs, effective and efficient DL architectures are crucial for the wide application of automated AV systems. Although there are significant improvements in 3D DL models, such as PointNet \cite{qi2017pointnet}, PointNet++ \cite{qi2017pointnet++}, PointCNN \cite{li2018pointcnn}, DGCNN \cite{wang2018dynamic}, RotationNet \cite{kanezaki2018rotationnet} and other work \cite{li2019discrete,liu2019relation,worrall2018cubenet,fujiwara2018canonical}. Some limited models can achieve real-time segmentation, detection and classification tasks. Researches should focus on lightweight and compact architecture designing. 

\subsubsection{\textbf{Context Knowledge Extraction}} 
 Due to the sparsity of point clouds and incompleteness of scanned objects, detailed context information for objects is not fully exploited. For example, the semantic contexts in traffic signs are crucial cues for AVs navigation, but existing deep models cannot extract such information completely from point clouds. Although multi-scale feature fusion approaches \cite{dong2018hierarchical,deng2018ppfnet,xie2018attentional} have demonstrated significant improvements in context information extraction. Besides, GAN \cite{wen2019deep} can be utilized to improve the completeness of 3D point clouds. However, these frameworks cannot solve the sparsity and incompleteness problems for context information extraction in an end-to-end trainable way. 

\subsubsection{\textbf{Multi-task Learning}} 
The approaches related to LiDAR point clouds for AVs consist of several tasks, such as scene segmentation, object detection (e.g., cars, pedestrians, traffic lights, etc.) and classification (e.g., road markings, traffic signs). All these results are commonly fused together and reported to a decision system for final control \cite{janai2017computer}. However, there are few DL architectures combining these multiple LiDAR point cloud tasks together \cite{qi2016volumetric,sedaghat2016orientation}. Thus, the inherent information among them is not fully exploited and used to generalize better models with less computation.

\subsubsection{\textbf{Weakly Supervised/Unsupervised Learning}} The existing state-of-art deep models are commonly constructed under supervised modes using labeled data with 3D objects bounding boxes or per-point segmentation masks \cite{wang2018dynamic,yang2018ipod,yang2018pixor}. However, there are some limitations for fully supervised models. First is the limited availability of high quality, large scale, and enormous general objects datasets and benchmarks. Second is the fully-supervised model generalization capability which is not robust to unseen or untrained objects. Weakly supervised \cite{yew20183dfeat} or unsupervised learning \cite{sauder2019context,shoef2019pointwise} should be developed to increase the model's generalization and solve the data absence problem.

\section{Conclusion}
In this paper, we have provided a systematic review of the state-of-the-art DL architectures using LiDAR point clouds in the field of autonomous driving for specific tasks such as segmentation, detection, and classification. Milestone 3D
deep models and 3D DL applications on these three tasks have been summarized and evaluated with merits and demerits comparison.
Research challenges and opportunities were listed to advance the potential development of DL in the field of autonomous driving.

\section*{Acknowledgment}
The authors would like to thank the Professors José Marcato Junior and Wesley Nunes Gonçalves for their carefully proofreading. Besides, we also would like to thank anonymous reviewers for their insightful comments and suggestions.

\ifCLASSOPTIONcaptionsoff
  \newpage
\fi


\bibliographystyle{IEEEtran}
\bibliography{mybibliography}

%



%
\vskip -2\baselineskip plus -1fil

\begin{IEEEbiography}[{\includegraphics[width=1in,height=1.25in,clip,keepaspectratio]{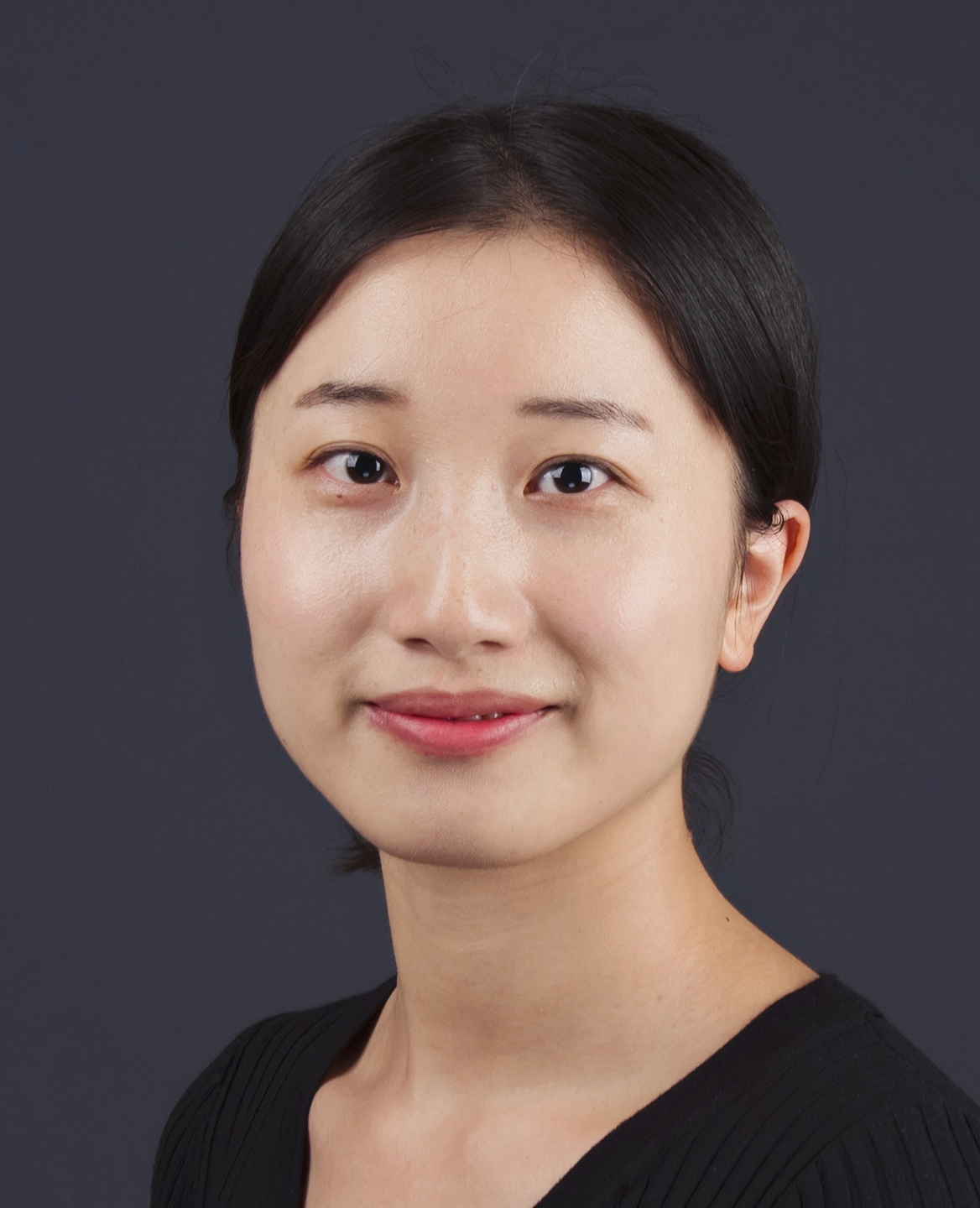}}]{Ying Li}
received the M.Sc. degree in remote sensing from Wuhan University, China, in 2017. She is currently working toward the Ph.D. degree with the Mobile Sensing and Geodata Science Laboratory, Department of Geography and Environmental Management, University of Waterloo, ON, Canada.

Her research interests include autonomous driving, mobile laser scanning, intelligent processing of point clouds, geometric and semantic modeling, and augmented reality.
\end{IEEEbiography}

\vskip -2\baselineskip plus -1fil

\begin{IEEEbiography}[{\includegraphics[width=1in,height=1.25in,clip,keepaspectratio]{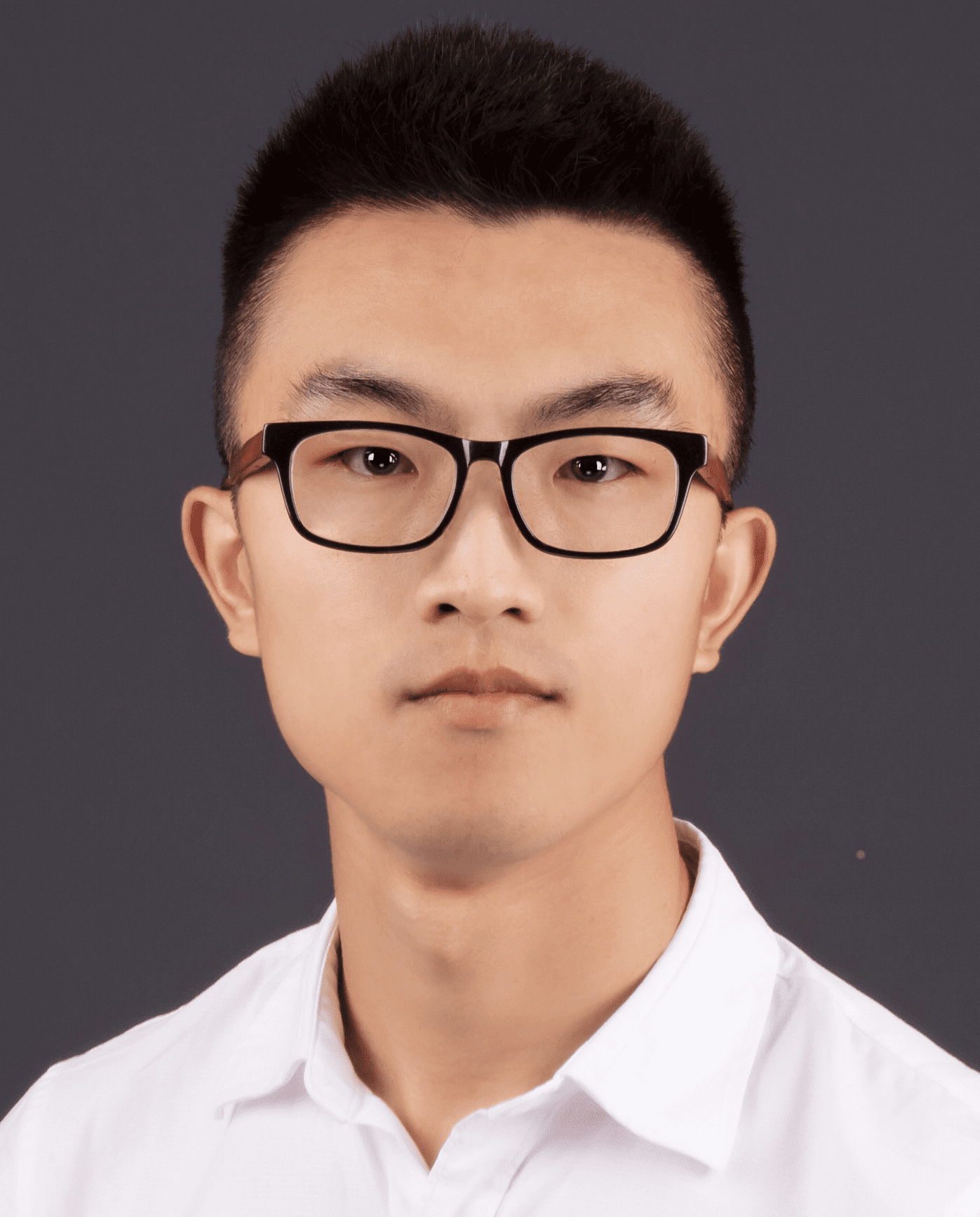}}]{Lingfei Ma}
(S’18) received the B.Sc. and M.Sc. degrees in geomatics engineering from the University of Waterloo, Waterloo, ON, Canada, in 2015 and 2017, respectively. He is currently working toward the Ph.D. degree in photogrammetry and remote sensing with the Mobile Sensing and Geodata Science Laboratory, Department of Geography and Environmental Management, University of Waterloo.

His research interests include autonomous driving, mobile laser scanning, intelligent processing of point clouds, 3-D scene modeling, and machine learning.
\end{IEEEbiography}

\vskip -2\baselineskip plus -1fil

\begin{IEEEbiography}[{\includegraphics[width=1in,height=1.25in,clip,keepaspectratio]{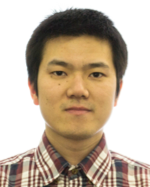}}]{Zilong Zhong}
(S’15) Zilong Zhong received the Ph.D. in systems design engineering, specialized in machine learning and intelligence, from the University of Waterloo, Canada, in 2019. He is a postdoctoral fellow with the School of Data and Computer Science, Sun Yat-Sen University, China. His research interests include computer vision, deep learning, graph models and their applications involving large-scale image analysis.
\end{IEEEbiography}

\vskip -2\baselineskip plus -1fil

\begin{IEEEbiography}[{\includegraphics[width=1in,height=1.25in,clip,keepaspectratio]{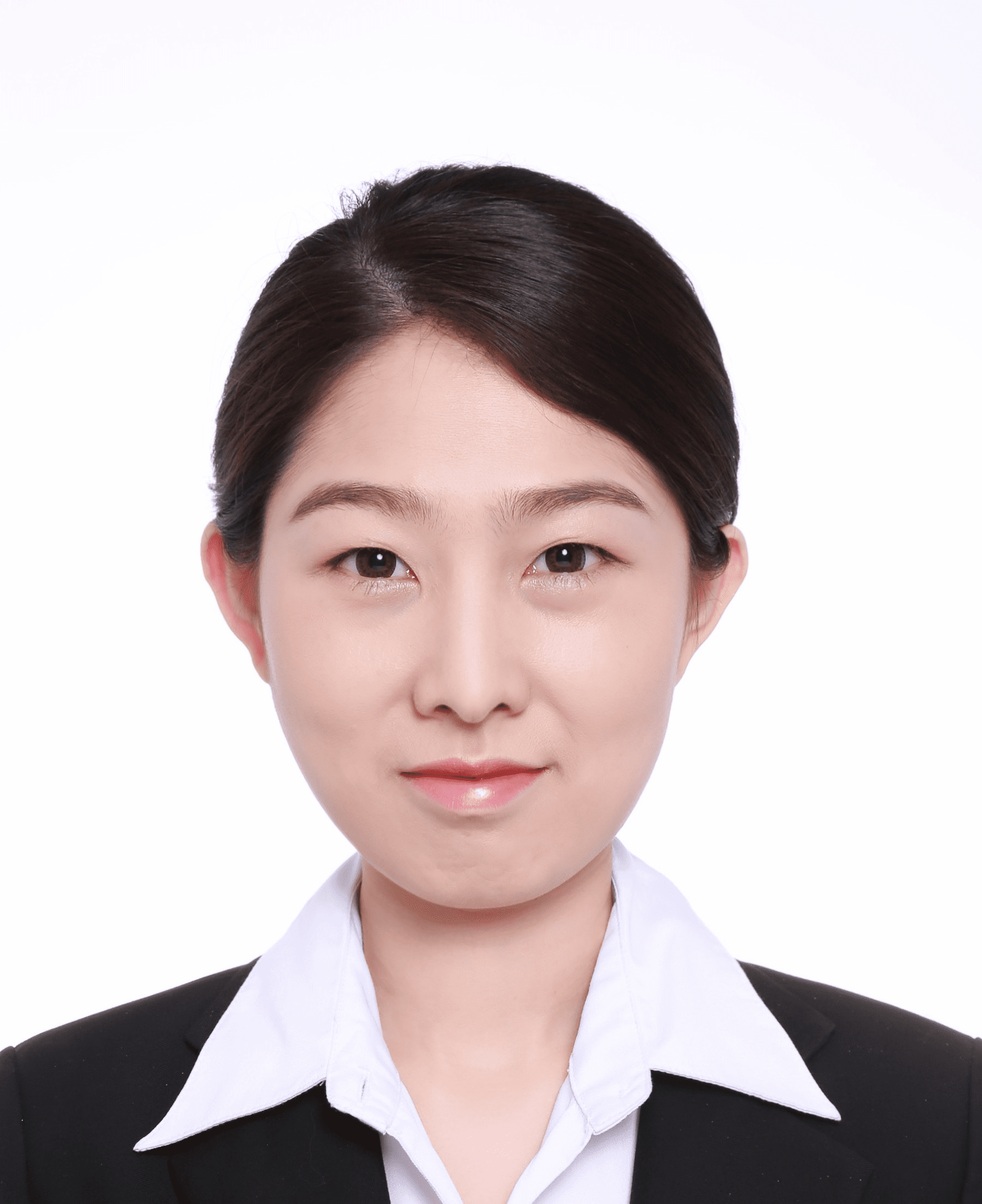}}]{Fei Liu}
received the B.Eng. degree from Yanshan University, China, in 2011. Since then, she has been working in the sectors of vehicular electronics and control, artificial intelligence, deep learning, FPGA, advanced driver assistance systems, and automated driving.

She is currently working at Xilinx Technology Beijing Limited, Beijing, China, focusing on development of automated driving technologies and data centers.  
\end{IEEEbiography}

\vskip -2\baselineskip plus -1fil

\begin{IEEEbiography}[{\includegraphics[width=1in,height=1.25in,clip,keepaspectratio]{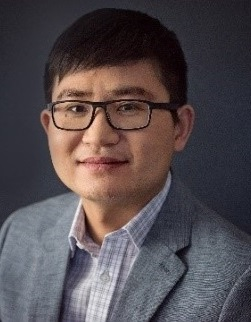}}]{Dongpu Cao}
 o (M’08) received the Ph.D. degree from Concordia University, Canada, in 2008. 
 
 He is the Canada Research Chair in Driver Cognition and Automated Driving, and currently an Associate Professor and Director of Waterloo Cognitive Autonomous Driving (CogDrive) Lab at University of Waterloo, Canada. His current research focuses on driver cognition, automated driving and cognitive autonomous driving. He has contributed more than 200 publications, 2 books and 1 patent. He received the SAE Arch T. Colwell Merit Award in 2012, and three Best Paper Awards from the ASME and IEEE conferences. 
\end{IEEEbiography}

\vskip -2\baselineskip plus -1fil

\begin{IEEEbiography}[{\includegraphics[width=1in,height=1.25in,clip,keepaspectratio]{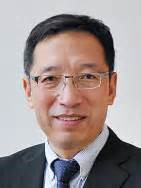}}]{Jonathan Li }
(00' M- 11' SM) received the Ph.D. degree in geomatics engineering from the University of Cape Town, South Africa. 

He is currently a Professor with the Departments of Geography and Environmental Management and Systems Design Engineering, University of Waterloo, Canada. He has coauthored more than 420 publications, more than 200 of which were published in refereed journals, including the IEEE Transactions on Geoscience and Remote Sensing, IEEE Transactions on Intelligent Transportation Systems, IEEE Journal of Selected Topics in Applied Earth Observations and Remote Sensing, ISPRS Journal of Photogrammetry and Remote Sensing, Remote Sensing of Environment, as well as leading artificial intelligence and remote sensing conferences including CVPR, AAAI, IJCAI, IGARSS and ISPRS. His research interests include information extraction from LiDAR point clouds and from earth observation images.
\end{IEEEbiography}

\vskip -2\baselineskip plus -1fil

\begin{IEEEbiography}[{\includegraphics[width=1in,height=1.25in,clip,keepaspectratio]{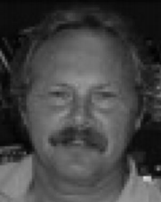}}]{Michael A. Chapman}
received the Ph.D. degree in photogrammetry from Laval University, Canada.

He was a Professor with the Department of Geomatics Engineering, University of Calgary, Canada, for 18 years. Currently, he is a Professor of geomatics engineering with the Department of Civil Engineering, Ryerson University, Canada. He has authored or coauthored over 200 technical articles. His research interests include algorithms
and processing methodologies for airborne sensors using GNSS/IMU, geometric processing of digital imagery in industrial
environments, terrestrial imaging systems for transportation infrastructure mapping, and algorithms and processing strategies for biometrology applications.
\end{IEEEbiography}




\end{document}